\documentclass[10pt,journal,letterpaper,compsoc]{IEEEtran}

\renewcommand\paragraph[1]{}

\usepackage{graphicx}
\usepackage{epsfig}
\usepackage[lofdepth,lotdepth]{subfig}
\usepackage{amsmath} %
\usepackage{amssymb}  %
\usepackage{hyperref}

\title{\LARGE \bf Cutaneous Force Feedback as \\ a Sensory Subtraction Technique in Haptics}

\author{Domenico Prattichizzo,~\IEEEmembership{Member,~IEEE,} Claudio Pacchierotti,~\IEEEmembership{Student Member,~IEEE,} and Giulio Rosati%
  \IEEEcompsocitemizethanks{\IEEEcompsocthanksitem{D. Prattichizzo and C. Pacchierotti are with the
      Department of Information Engineering, University of Siena, Via Roma 56, Siena, Italy and with the
      Department of Advanced Robotics, Istituto Italiano di Tecnologia, Via Morego 30, Genova, Italy. G. Rosati is
      with the Department of Management and Engineering, University of Padua, Padua, Italy.}
  }%
}

\begin{document}

\IEEEcompsoctitleabstractindextext{%
\begin{abstract}
A novel sensory substitution technique is presented.  Kinesthetic and cutaneous force feedback are
substituted by cutaneous feedback (CF) only, provided by two wearable devices able to apply forces to
the index finger and the thumb, while holding a handle during a teleoperation task.  The force
pattern, fed back to the user while using the cutaneous devices, is similar, in terms of intensity
and area of application, to the cutaneous force pattern applied to the finger pad while
interacting with a haptic device providing both cutaneous and kinesthetic force feedback. The
pattern generated using the cutaneous devices can be thought as a subtraction between the complete
haptic feedback (HF) and the kinesthetic part of it. For this reason, we refer to this approach as
\textit{sensory subtraction} instead of sensory substitution.
A needle insertion scenario is considered to validate the approach.  The haptic device is
connected to a virtual environment simulating a needle insertion task.  Experiments show that the
perception of inserting a needle using the cutaneous-only force feedback is nearly indistinguishable
from the one felt by the user while using both cutaneous and kinesthetic feedback.  As most of the
sensory substitution approaches, the proposed sensory subtraction technique also has the advantage
of not suffering from stability issues of teleoperation systems due, for instance, to
communication delays. Moreover, experiments show that the sensory subtraction technique
outperforms sensory substitution with more conventional visual feedback (VF).
\end{abstract}

}

\maketitle

\section{Introduction} %

\paragraph{\footnotesize\bf [Summary: Sensory substitution and medical
  applications]}
A novel approach to sensory substitution in haptics is presented.  Sensory substitution is used in
teleoperation to display forces using other modalities such as audio or visual feedback (VF) or other
forms of haptic feedback (HF) such as vibrotactile feedback.  Sensory substitution techniques are
frequently used in medical applications \cite{okamura2009}. In this paper we focus on a simulated
environment for teleoperated needle insertion in soft tissues.

\paragraph{\footnotesize\bf [Summary: Needle insertion]}
In recent years studies on needle insertion in soft tissues have
attracted considerable attention due to their promising applications
in minimally invasive percutaneous procedures such as biopsies
\cite{bishoff1998}, blood sampling \cite{zivanovic2000}, neurosurgery
\cite{masamune1995} \cite{rizun2004}, and brachytherapy
\cite{dimaio2003} \cite{hing2006}.
The effectiveness of a treatment depends on the accuracy of
percutaneous insertion \cite{abolhassani2007} \cite{hing2006},
especially when working on critical areas like the brain.

Force feedback is an important navigation tool during surgical needle
advancement. It allows to detect local mechanical properties of the
tissue being penetrated and distinguish between expected and abnormal
resistance due, for example, to the unexpected presence of vessels
\cite{lorenzoforce}.  An interesting study on the effect of
teleoperation on perception abilities of human operators on the
stiffness of the tissue has been recently presented in
\cite{Nisky2011}.

\paragraph{\footnotesize\bf [Summary: Problems: stability and
  transparency]}
In bilateral teleoperation, stability and transparency can be
significantly affected by communication latency of the teleoperation
loop which dramatically reduces the effectiveness of haptic feedback
in case of stiff remote environments
\cite{lawrence1993} \cite{hashtrudi2002}.
This limitation can be alleviated requiring passivity of the interconnected system \cite{lee2006},
using wave variable transformation \cite{niemeyer1991} \cite{ye2009} or designing proper control
systems \cite{anderson1989} \cite{adams1998}.
\paragraph{\footnotesize\bf [Summary: the main issue of algorithmic
  solutions ]}
However, designing proper control algorithms to guarantee stability cannot be considered as an
intrinsically safe approach. To prevent serious mechanical faults such as actuator failures on the
master side, which can generate undesired and unsafe motions of the slave robot, different
approaches must be considered. In particular, we need to consider techniques dealing more with the
hardware design than the control architecture of the teleoperation loop.
\paragraph{\footnotesize\bf [Summary: Hardware solutions]}

\begin{figure}[!t]
\begin{centering}
\includegraphics[height=120pt]{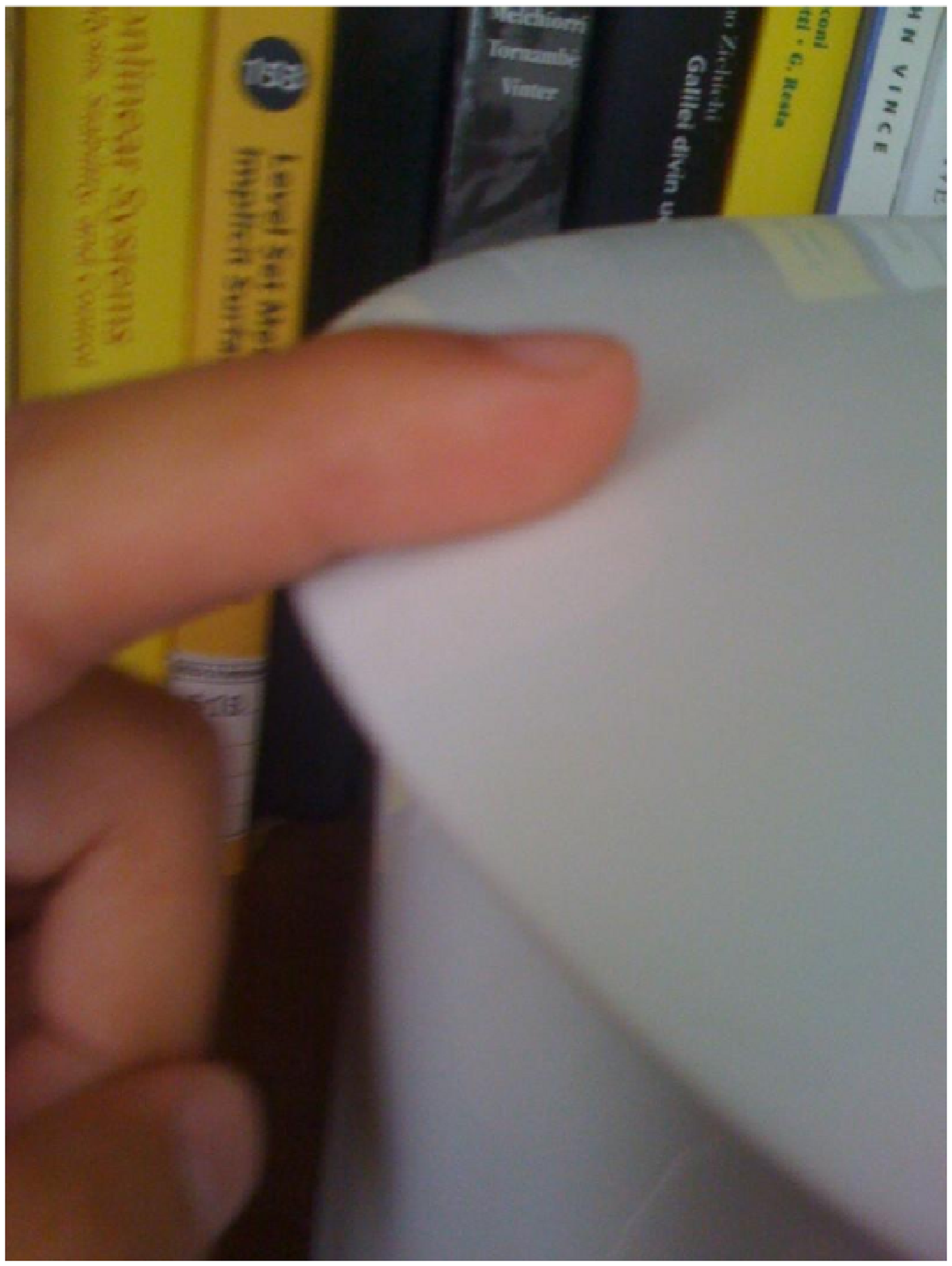}
\includegraphics[height=120pt]{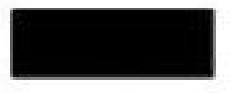}
\includegraphics[height=120pt]{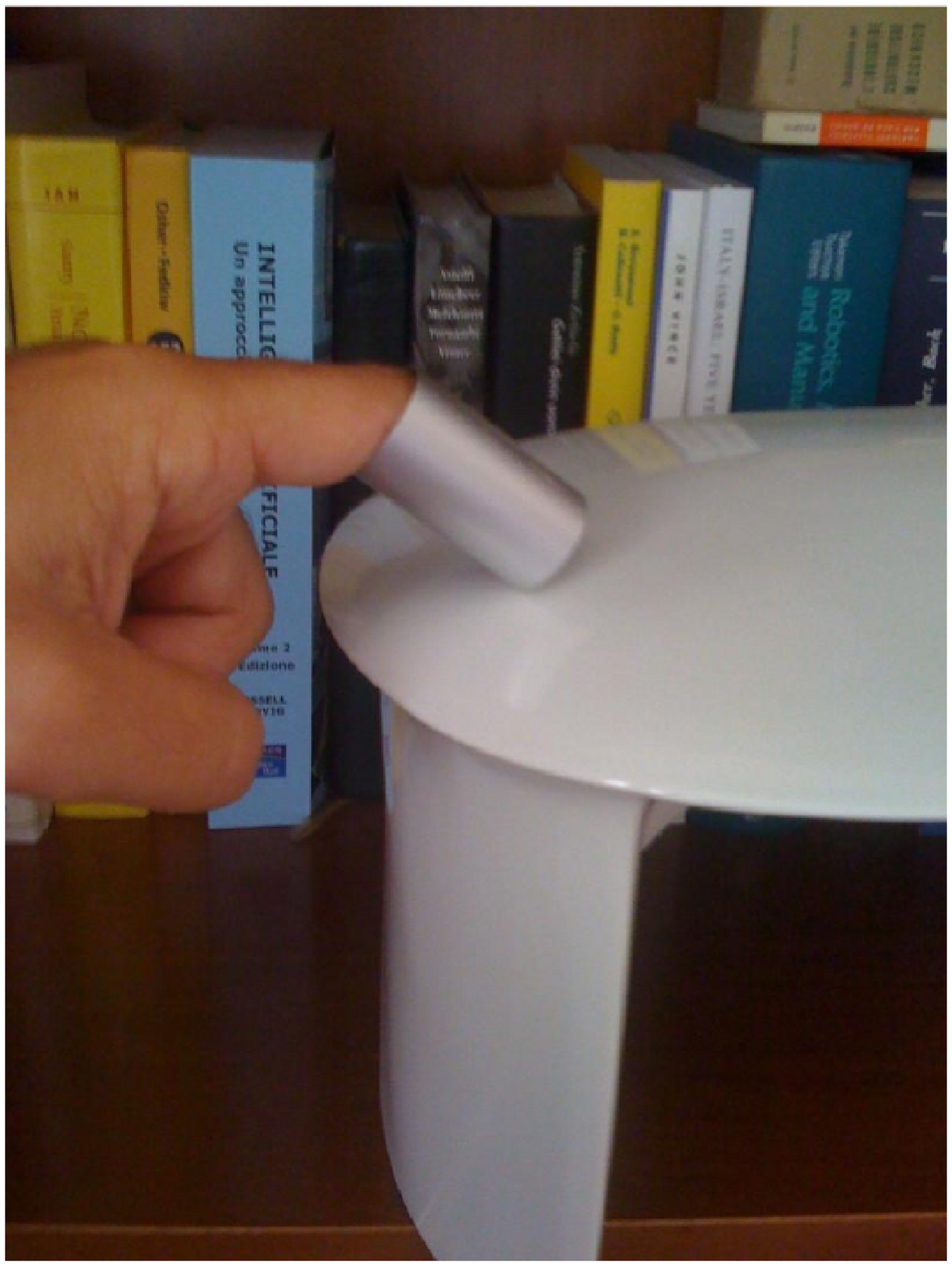}
\caption{While touching an object, the human feels both kinesthetic
  and cutaneous stimuli (left); isolating the fingertip skin with a
  thimble makes the user perceive kinesthetic interaction mainly
  (right). In the interaction with haptic devices, the subtraction of kinesthetic feedback from the mixed stimuli
  brings to cutaneous only feedback.}
\label{subtraction}
\end{centering}
\end{figure}

In the literature, a possible hardware design approach consists in using passive components such as
brakes \cite{black2009} or passive isometric input devices \cite{lecuyer2000}.  However, passive
input devices have rendering limitations and may lead to large steady-state errors in teleoperation
tasks.
To reduce the effects of these limitations,  researchers implemented
energy-bounding algorithms \cite{kim2010} or used motors and brakes
together with the aim of obtaining a safer teleoperation while
preserving system transparency \cite{conti2009}.

\paragraph{\footnotesize\bf [Summary: Substitution]}
Another interesting approach consists in avoiding to use any actuator for force feedback on the
master side and alternatively providing the force feedback using sensory substitution techniques.
Force feedback is not kinesthetic anymore and the haptic loop becomes intrinsically stable since no
force is fed back to the operator through the haptic device.  Sensory substitution techniques
replaces this lack of kinesthetic feedback with other forms of feedback such as vibrotactile
\cite{schoonmaker2006} \cite{massimino1993}, %
auditory, and/or visual feedback \cite{tavakoli2005} \cite{kitagawa2005}.

\paragraph{\footnotesize\bf [Summary: Our contribution]}
The sensory subtraction technique presented in this work can be casted in a sensory substitution
framework but there are relevant differences which are worth underlining to motivate the use of the
term {\em subtraction}.  The main idea is that, instead of rendering forces with a complete haptic
feedback, consisting of cutaneous and kinesthetic components, we present to the human operator the
cutaneous component only, without the kinesthetic part.  A novel wearable cutaneous force feedback
device has been developed for this aim. Differently from other works on cutaneous feedback (CF), the
device presented in this paper is not of the array type as discussed, for instance, in
\cite{Wagner02} and \cite{Garcia2011}, but it allows to apply vertical stresses to the finger pad,
similarly to the gravity grabber presented in \cite{minamizawa2007gravity}.  The role of cutaneous
feedback in haptics, compared to kinesthetic feedback, has been recently discussed and exploited,
for example, in \cite{Wijntjes09}, where Wijntjes et al. discussed the effects of kinesthetic and
cutaneous information for curvature discrimination, in \cite{Ferber09}, where Ferber et al.
investigated cutaneous and kinesthetic cues to maintain exercise intensity on a stair climber
machine and in \cite{MiPrTa10-hs10} where the problem of missed kinesthetic feedback in wearable
haptics is discussed. All these papers underline how relevant is the cutaneous feedback when
compared to kinesthesia.

In this work, we will show how the proposed cutaneous-feedback sensory subtraction technique, other
than being intrinsically stable, improves the teleoperation performances with respect to other
sensory substitution techniques such as the one using visual feedback.  Preliminary results on the
sensory subtraction approach were presented for an industrial application in
\cite{prattichizzo2010using}, where the the gravity grabber presented in
\cite{minamizawa2007gravity} was used. In this paper, we discuss the results of experiments on
sensory subtraction for needle insertion in which we use new wearable cutaneous devices.

\paragraph{\footnotesize\bf [Summary: The structure of the paper]}
The rest of the paper is organized as follows: the idea of sensory subtraction is discussed in
Sec. \ref{tactile} along with the description of the cutaneous device. The teleoperated needle
insertion application is introduced in Sec. \ref{med}. Experiments carried out to validate the
proposed sensory subtraction technique are presented and discussed in Sec.
\ref{results}. %
Finally, Sec. \ref{conclusions} addresses concluding remarks and perspectives of the work.

\section{Sensory subtraction} %
\label{tactile}

The idea behind sensory subtraction originates from the observation that the stimuli received by the
user while holding a haptic handle consists of a cutaneous and a kinesthetic component.  Cutaneous
sensation is produced by pressure receptors in the skin and they are useful to recognize the local
properties of objects such as shape, edges, embossings and recessed features, thanks to a direct
measure of the intensity and direction of the contact forces \cite{Johansson2001}.  On the other
hand, kinesthesia provides the user with information about the relative position of neighboring
parts of the body, by means of sensory organs in the muscles and joints \cite{hayward2004}.

In this work, we propose to use an interface able to generate cutaneous force feedback only instead
of the complete haptic feedback (both kinesthetic and cutaneous), during the execution of a simple
robot-assisted surgical task. In particular, we will substitute the haptic force feedback with its
cutaneous component provided by a device able to apply normal forces to the finger pad.  With
respect to traditional haptic feedback, we expect this simple form of feedback to make the
teleoperation stable, and to allow the operator to perform the motion task in an equally intuitive
way, as the cutaneous force feedback is perceived where it is expected and provides the operator with a
direct and colocated perception of the contact force even if it is only cutaneous and not
kinesthetic.

Of course, when we use the cutaneous force feedback devices, the kinesthetic sensation is still
present because the hand and the arm move but not because a kinesthetic force feedback device is
acting on the user.  In other words, using a cutaneous force feedback device, we want to remove the
kinesthetic feedback produced by the actuators of the haptic device, rather than eliminating the
kinesthetic interaction.

We expect the proposed feedback modality to yield better results, in terms of task performance, with
respect to other forms of sensory substitution. For this reason, the novel feedback modality will be
compared not only to haptic feedback, but also to a common sensory substitution
technique, in which force feedback is substituted by a visual representation of the contact
force.

We could refer to our approach as sensory substitution because the
mixed kinesthetic-cutaneous feedback usually provided by a haptic device
is here {substituted} with part of the cutaneous feedback.  However, it
is worth underlining that here the stimulation fed back to the user is
similar, in terms of intensity and area of application, to the one
perceived while interacting with an actuated handle.  This approach is
different from other sensory substitution techniques in which the area
and/or the type of stimuli are different from the ones being replaced.

By considering that the handle of a haptic device would provide both kinesthetic and cutaneous force
feedback, and that the area where the force is applied is equivalent (i.e., the finger pad), using
our approach the user receives a subset of the typical stimuli provided by a haptic device. This is
why we refer to the proposed approach as \textit{sensory subtraction} (see Fig.~\ref{subtraction}).

\subsection{The wearable fingertip cutaneous force feedback device}

\begin{figure}[!t]
\begin{centering}
\includegraphics[height=110px]{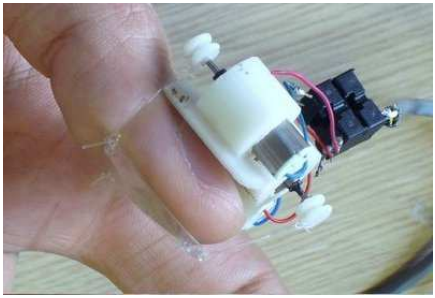}
\caption{The wearable cutaneous device used to apply forces normal to the
  operator's finger pad.
  }
\label{device}
\end{centering}
\end{figure}

\begin{figure}[b]
\begin{centering}
\includegraphics[height=110px]{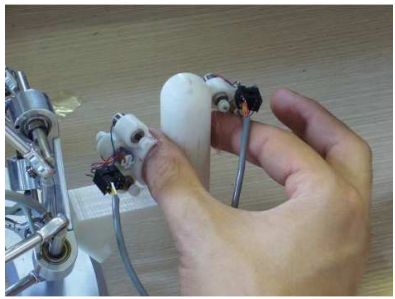}
\caption{The haptic handle grabbed by the operator using two fingers
  and two wearable cutaneous devices.}
\label{fingers}
\end{centering}
\end{figure}

The prototype of the cutaneous force feedback device used in our experiments is shown in
Fig. \ref{device}. It consists of two main parts: the first one is on the dorsal side of the finger
and supports three small electrical motors; the other has a contact patch with the volar skin
surface of the fingertip. The two parts are connected by three cables.  The motors, by controlling
the lengths of the cables, are able to press the patch on the user's fingertip. As a result, a force
is generated simulating the contact of the fingertip with the surface of an object or a handle, as
in Fig.~\ref{fingers}. The direction and amount of the force reflected to the user is changed by
properly controlling cable lengths \cite{ChMaPaPr-hs12}.

This device applies forces between the volar skin surface and the nail. In contrast, when humans
actively exert fingertip forces during manipulation of real objects, forces operate essentially
between the phalangeal bone and the volar skin surface.  Birznieks {\em et al.} \cite{Johansson2001}
demonstrated that the deformational changes in the fingertip are similar under the two conditions,
i.e., when stimulated by a device similar to the one proposed in this work, the fingertip will
deform as if the subject was actively applying forces against a real object. For this reason, the
cutaneous stimulation produced by the wearable device can be considered to some extent equivalent to
that perceived while actively interacting with a haptic handle (Fig.~\ref{fingers}).

The device described above belongs to the category of wearable haptic devices and it is an evolution
of the first idea presented by K. Minamizawa {\em et al.} \cite{minamizawa2007gravity}. In
particular, the evolution consists of using three motors instead of two, and a $3$-dof parallel
manipulator architecture \cite{Dasgupta200015} to render forces at the finger pad.  For the purpose
of this work, the wearable device in Fig.~\ref{device} was controlled as a $1$-dof system (all
motors pulled the cables together), so that only forces in the sagittal plane of the finger were
actuated, roughly normal to the longitudinal axis of the distal phalanx.

\subsection{Sensory subtraction - a demonstrator}

\begin{figure}[]
\begin{centering}
\includegraphics[height=110px]{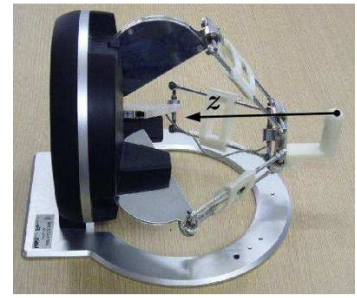}
\caption{The haptic device Omega 3, with three clamps limiting the
  motion of the handle along the $z$--axis only.}
\label{omega}
\end{centering}
\end{figure}

In our experiments, we used four prototypes of the fingertip cutaneous device and a commercial
haptic device.  The operator wears two cutaneous devices on one hand, one on the thumb and one on
the index finger, and grabs the handle as shown in Fig.~\ref{fingers}.  Two additional cutaneous
devices are worn on the thumb and index finger of the contralateral hand.  The haptic device is the
Omega 3 by Force Dimension, to which three clamps were applied to reduce the degrees of freedom from
three to one (the $z$ axis in Fig. \ref{omega}). Also, a plastic handle was attached to its
end-effector to allow the operator to grab the device with two fingers (Fig.~\ref{fingers}).

During the experiments, the hardware was operated in two different modalities. The first one is
referred to as \emph{complete haptic feedback}, where the feedback force is provided by the Omega 3
while the wearable devices are switched off. In this way, by interacting with the handle, the
operator receives mixed kinesthetic and cutaneous stimuli, i.e., the complete haptic feedback.

The second modality is referred to as \emph{cutaneous--only feedback}, where the proposed sensory
subtraction technique is implemented.  In this modality, the Omega 3 is used only to track the
motion of the hand with its encoders and does not apply any active force to the operator (the
actuators of the Omega 3 are switched off). At the same time, the wearable devices are used to
reproduce the cutaneous sensation associated to the manipulation task being simulated.  For
instance, a feedback force directed towards the negative direction of the $z$-axis (see
Fig. \ref{omega}) is substituted by applying a normal stress to the index finger. Conversely, a
force directed towards the positive direction of the $z$-axis is substituted by a normal stress
applied to the thumb. To investigate the role of feedback localization with respect to the hand
involved in the task, either the devices on the active hand or those worn on the contralateral hand
are alternatively activated.

\section{A medical application of sensory subtraction}
\label{med}

In this work, we test the sensory subtraction approach on a simulated scenario of needle insertion
in a soft tissue. Force feedback is helpful during needle advancement to
detect local mechanical properties of the tissue and to distinguish between expected and abnormal
resistance due, for example, to the unexpected presence of vessels, or to the action of active
constraints, that are usually introduced to protect areas of the soft tissue that must be avoided to
prevent damage of tissue and of its functionality. This is the case, for instance, of brain surgery,
in which tissue manipulation in special areas can cause serious injury to patients.

Active constraints, commonly referred to as virtual fixtures
\cite{rosenberg1993}, are software functions used in assistive robotic
systems to regulate the motion of surgical implements. The motion of
the surgical implement, the needle in our case, is still controlled by
the surgeon, but the system constantly monitors its motion and takes
some actions if the surgical tool fails to follow a predetermined procedure.
Virtual fixtures play two main roles: they can either guide the motion
or strictly forbid the surgeon from reaching certain regions
\cite{abbott2007}. A guiding virtual fixture attenuates the
motion of the surgical implement in some predefined directions to
encourage the surgeon to conform to the procedure plan.  A
forbidden-region virtual fixture is a software constraint that seeks
preventing the needle from entering a specific region of the workspace.
In this paper, we consider an example of virtual fixtures protecting
forbidden regions. This is a common scenario for biopsies, deep brain
stimulation and functional neurosurgery.%

When performing keyhole neurosurgery the needle can be steered using a haptic device such as the
Omega~3, and the motion of the needle will be along one direction only \cite{lorenzoforce,davies2003}.  The device used in the experiments is reported in Fig. \ref{fingers}. A special
handle is attached to the end-effector and the motion is constrained to one degree of freedom, by
means of three clamps attached to the parallel structure of the device.  The Omega 3 is typically
used as a haptic device of the impedance type: the position of the needle, moved by the human
operator, is measured, and a force signal is fed back to the user through the actuation system.  The
force feedback accounts for either the remote contact interaction of the slave robot, in a classical
teleoperation scenario, or by the virtual environment, in case of simulations.

\begin{figure}[t]
\begin{centering}
 \subfloat[][no needle-tissue contact]{
 \label{ve:1}
\includegraphics[width=135pt]{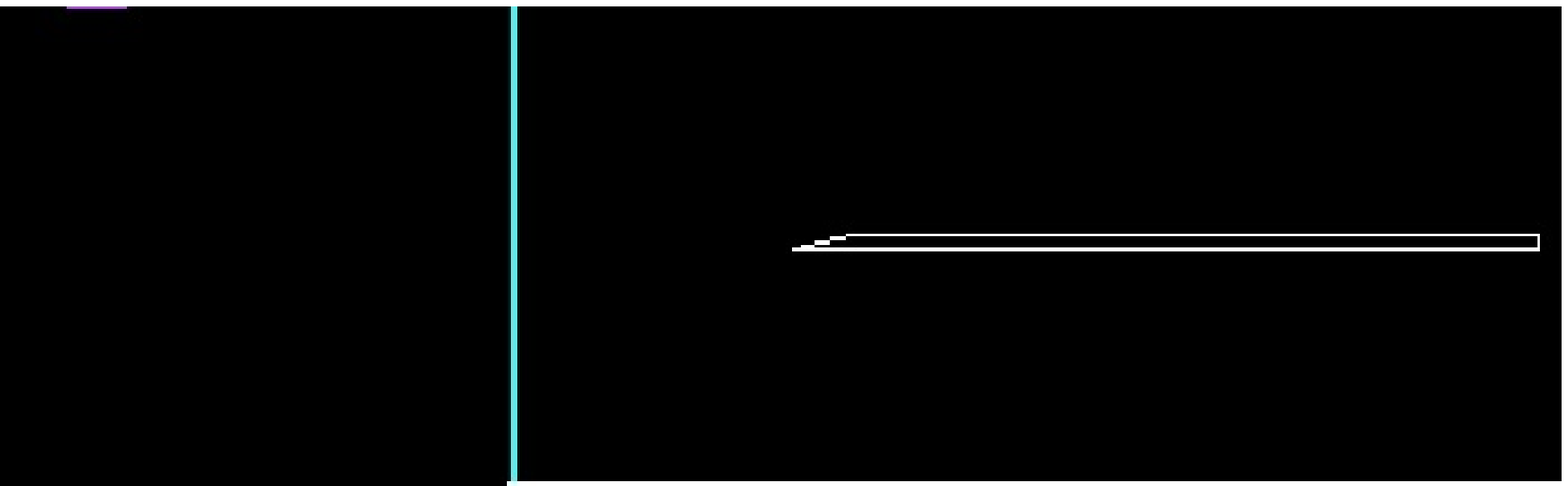}
}

 \subfloat[][needle reaches the tissue]{
 \label{ve:2}
\includegraphics[width=135pt]{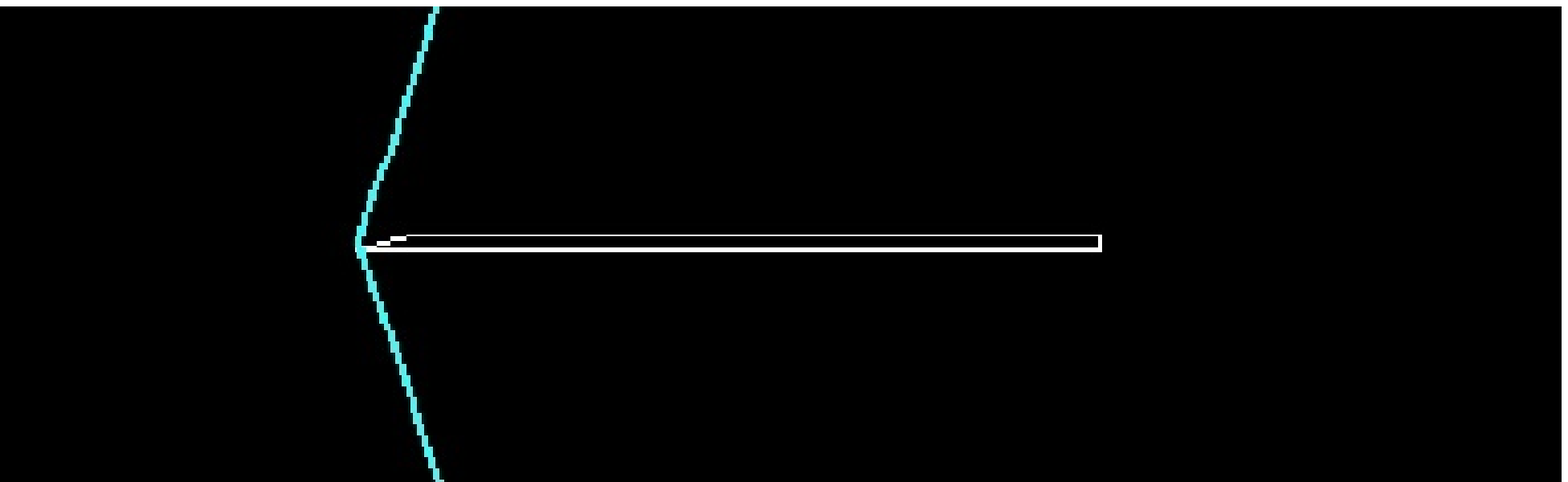}
}

 \subfloat[][needle penetrates the tissue]{
 \label{ve:3}
\includegraphics[width=135pt]{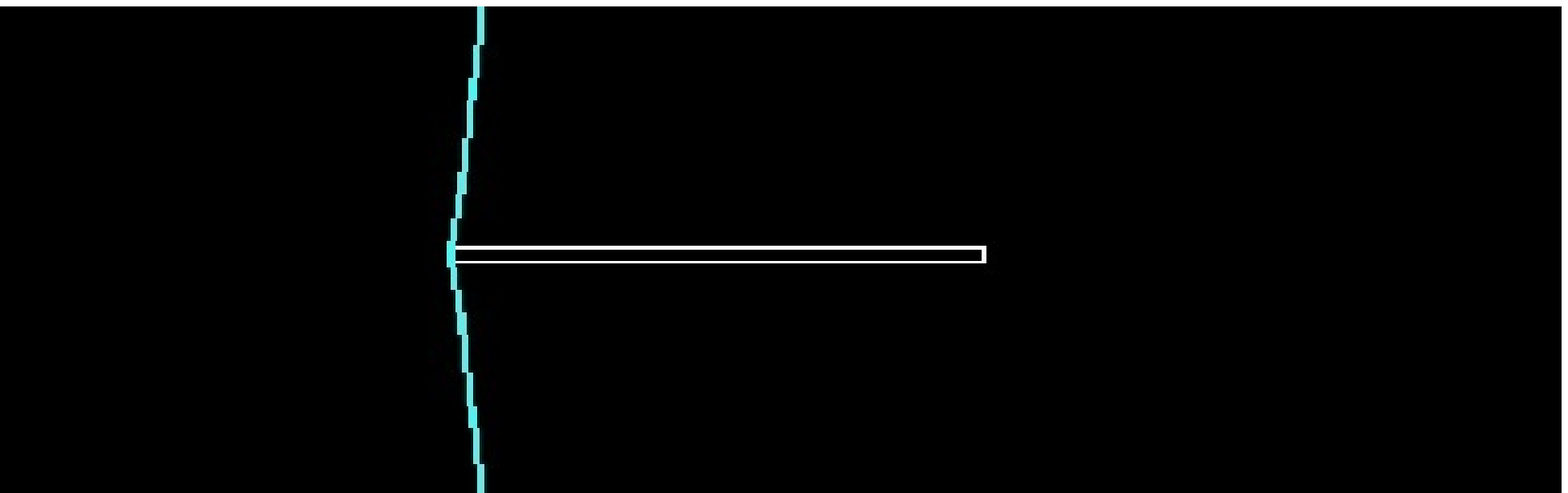}
}
\caption{Screenshots of the virtual environment, composed by the needle (white),
  driven by the operator, and the deformable tissue (light blue).
  The portion of the needle already inserted in the tissue is not shown on screen (c).}
\label{ve}
\end{centering}
\end{figure}

In the proposed setup, the haptic handle teleoperates the needle in a
virtual environment simulating the insertion in a soft tissue with
virtual fixtures.  The needle moves along a single axis (the
$z$-axis of the haptic device) as in Fig. \ref{ve}, where the needle
and the surface of the tissue are shown. The contact force between the
needle and the tissue is calculated according to the visco-elastic
model presented in Sec. \ref{tissue}.

\subsection{Soft tissue modeling and haptic rendering}
\label{tissue}

The operator remotely steering
the needle feels a resistive force, while penetrating the tissue, due to its visco-elastic properties, and an opposite force while trying to pull the needle out.  In real
scenarios, these forces are either measured from force sensors or
estimated from other parameters. 

In this work a simple simulation of the soft tissue is used. The aim
of this work is not to design an accurate tissue simulator based for
instance on FEM techniques \cite{cotin1999} but to validate the proposed sensory subtraction approach.

A spring $K_{t}= 2$~N/m and a damper $B_{t} = 5$~Ns/m are used to model the contact force $F_{t}$ between the needle and
the tissue, while a spring $K_{vf} = 3000$~N/m is used
to model the contact force $F_{vf}$ between the needle and the virtual fixture.
For the sake of simplicity, we assume that the mass of the
tissue $M_{t} = 1$~kg is concentrated at the contact point.  The
viscous coefficient of the body beneath the tissue is $V_{t}
= 0.7$~Ns/m.

As for the haptic rendering, the interaction is designed according
to the god-object model \cite{zilles1995} and the position of the Omega
handle is linked to the needle position $z_{n}$ moving in
the virtual environment.  The initial position of the surface
of the tissue is set to $\bar{z}_{t}= 20$~mm and the virtual fixture is located at $\bar{z}_{vf}= 123$~mm.

Tissue position $z_{t}$ changes according to the interaction
with the needle, which is able to penetrate the surface only when the
haptic force $F_{h}$ is larger than a predetermined threshold ($F_{p}= 0.1$~N).  To have a wider workspace, a scale factor of $3$ between the position of the needle in the virtual
environment and the operator's hand is used.

It is possible to discriminate four different operating
conditions for the needle-tissue interaction model here presented:
\begin{itemize}
\item no contact (see Fig. \ref{ve:1}),
\item contact without penetration (see Fig. \ref{ve:2}),
\item penetration within the safe area (see Fig. \ref{ve:3}), and
\item penetration and contact with the virtual fixture
\end{itemize} 
In the first case, since the needle is out of the tissue, the model is
designed to feed back no force to the operator and the surface of the tissue tends to return to its predetermined initial position $\bar{z}_{t}$.
The dynamics of the interaction for the no contact case is
\begin{displaymath}
\begin{cases}
M_{t}~\ddot{z}_{t} = -K_{t}~(z_{t}-\bar{z}_t) - B_{t}~\dot{z}_{t},\\
F_{h} = 0.
\end{cases}
\end{displaymath}
When the needle touches the tissue, but the force $F_{h}$ is not yet
sufficient to penetrate it, the tissue surface is deformed by
the movement of the needle. In this case, the dynamic model and the
contact force to be fed back to the operator are
\begin{displaymath}
\begin{cases}
  z_{t} = z_{n},\\
  F_{h} = - K_{t}~(z_{t} - \bar{z}_{t}) - B_{t}~\dot{z}_{t}.
\end{cases}
\end{displaymath}
As soon as $F_{h}>F_{p}$, the needle
penetrates the surface and while the needle is inside the tissue, the
dynamics and the contact force are computed as
\begin{displaymath}
\begin{cases}
M_{t}~\ddot{z}_{t} = -K_{t}~(z_{t}-\bar{z}_t) - B_{t}~\dot{z}_{t} - V_{t}~(\dot{z}_{t} - \dot{z}_{n}),\\
F_{h} = - V_{t}~(\dot{z}_{t}-\dot{z}_{n}).
\end{cases}
\end{displaymath}
If the operator steers the needle towards the unsafe workspace area delimited by the virtual fixture, a force will be fed back to the operator in order to avoid the penetration of the needle in the forbidden area:
\begin{center}$F_{vf} = - K_{vf}~(z_{n}-\bar{z}_{vf}).$\end{center}
Note that the virtual fixture generates a force feedback which is
more than $10^3$ times larger than the force felt while in contact with the soft tissue.

The haptic device measures the position of the operator's hand (with a resolution of $ 0.01$~mm),
sends it to the controller and then the virtual environment computes the force feedback and the
dynamics of the tissue.  The controller then sends the force back to the user through either the
haptic device or the substitutive (cutaneous or visual) modality.

\subsection{Design of experiments}
Four alternative feedback modalities were compared in the experiments: (complete) \textit{haptic}
feedback, applied by the actuators of the haptic handle, \textit{visual} feedback in substitution of
haptic feedback, \textit{cutaneous-only} feedback in substitution of haptic feedback, applied by the
wearable devices either on the fingers holding the handle \emph{or} on the fingers of the
contralateral hand.  The visual feedback consisted in showing a horizontal bar representing the
contact force registered at the needlepoint.%

The subjects were asked to wear the four cutaneous devices for the whole duration
of the experiments, and to grasp the handle with their right hand as shown in Fig.~\ref{fingers}.
The subject's hand was positioned with its longitudinal axis at $ 90 $ degrees from the Omega $z$-axis.
The position of the subject's hand with respect to the joystick was checked before the beginning of each experiment. 
To prevent changes in the perceived direction of the feedback force generated by the Omega 3, the subjects were instructed to move the forearm rather than the wrist while moving the device.
During the experiments, the subjects maintained the initial orientation of the fingers with respect to the handle, which was the only natural way of grasping the handle for the 1-dof task\footnote{A modification of the way the fingers grasp the handle would imply that the perceived direction of the feedback force changes if haptic feedback is used, whereas it would not change with cutaneous-only feedback. This issue must be considered while trying to extend the sensory subtraction paradigm to multi-dof tasks, since the results may be affected by this change of direction of the force vector. Thus, the position of the operator's hand with respect to the input device must be carefully monitored before and during the experiments.}.

The task consisted in inserting the needle into the soft tissue and
stopping the motion when a virtual fixture was perceived.
After $5$~s of continuous contact with the fixture, the system played a sound beep.
The subjects were instructed to pull the needle out of the tissue as soon as the sound was heard.
In all the considered tasks, regardless of particular feedback modality employed,
visual feedback on needle insertion was provided to the subjects, showing the part
of the needle out of the tissue and the surface of the tissue:
the virtual fixture and the portion of the needle inside the tissue
were not visible (see Fig.~\ref{ve}).%

No information on the feedback modalities was provided, neither on
their nature (except from visual feedback in substitution of force
feedback) nor on the particular order with which they were going to be presented
to the subject. Both the sequence of the feedback modalities and the
position of the virtual fixture were randomized.

\begin{figure}[]
\centering
	 \subfloat[][Visual feedback (VF)]{
		\includegraphics[width=0.85\columnwidth]{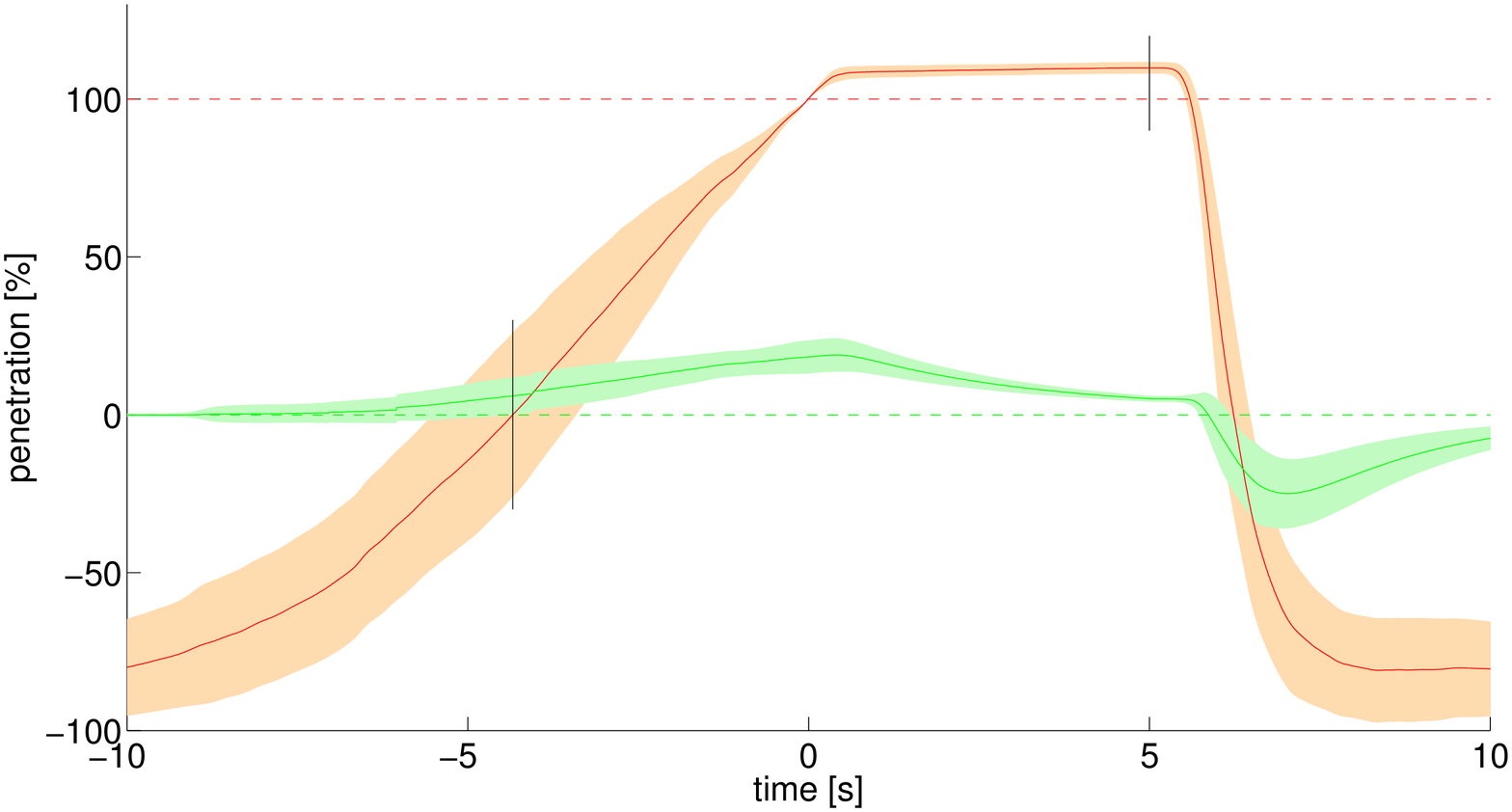}
		\label{nodelay:1}
	 }	 
	 
	 \subfloat[][Haptic feedback (HF)]{
		\includegraphics[width=0.85\columnwidth]{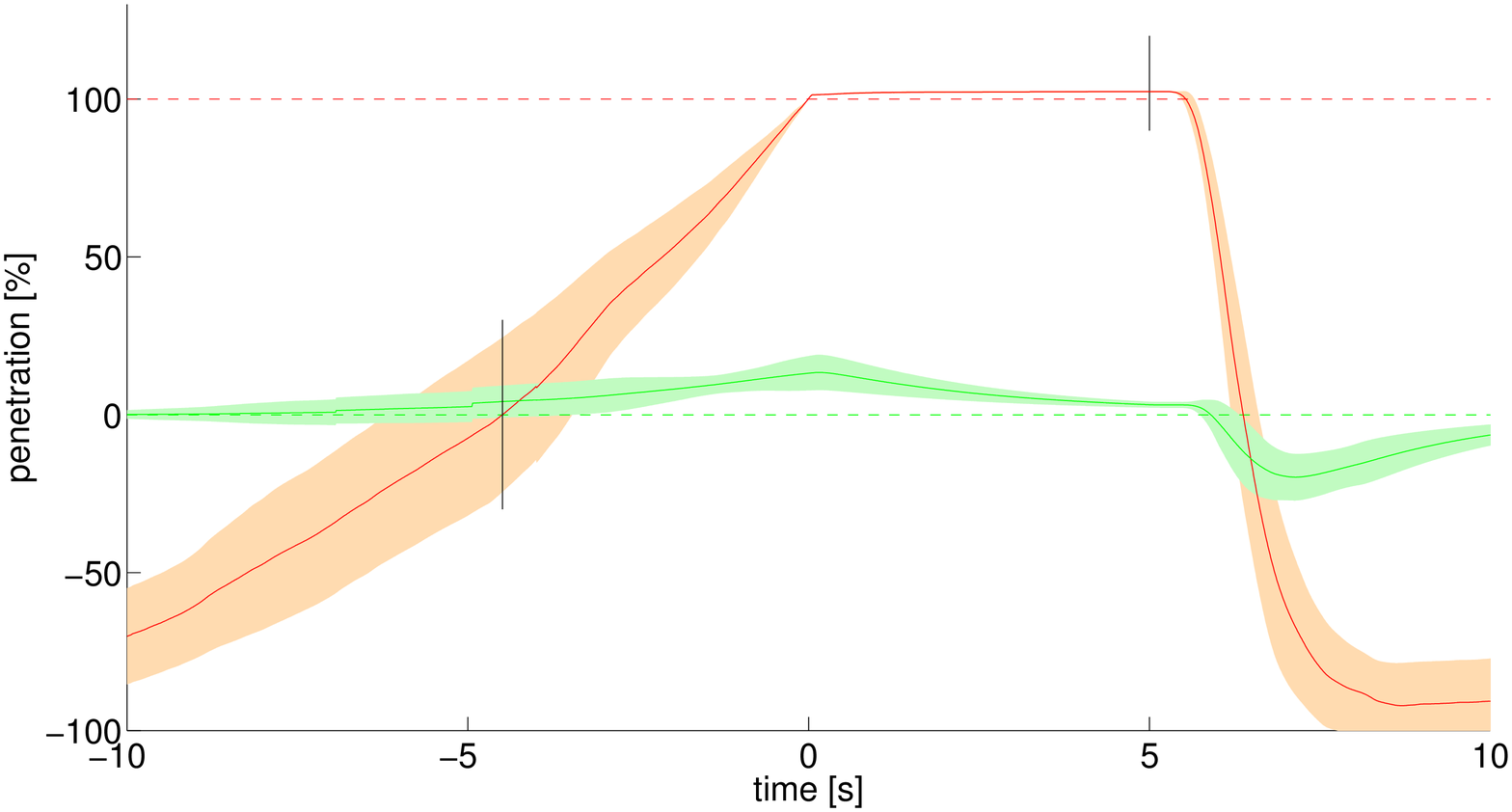}
		\label{nodelay:2}
	 }	 
	 
	 \subfloat[][Cutaneous feedback on the hand holding the handle (CF)]{
		\includegraphics[width=0.85\columnwidth]{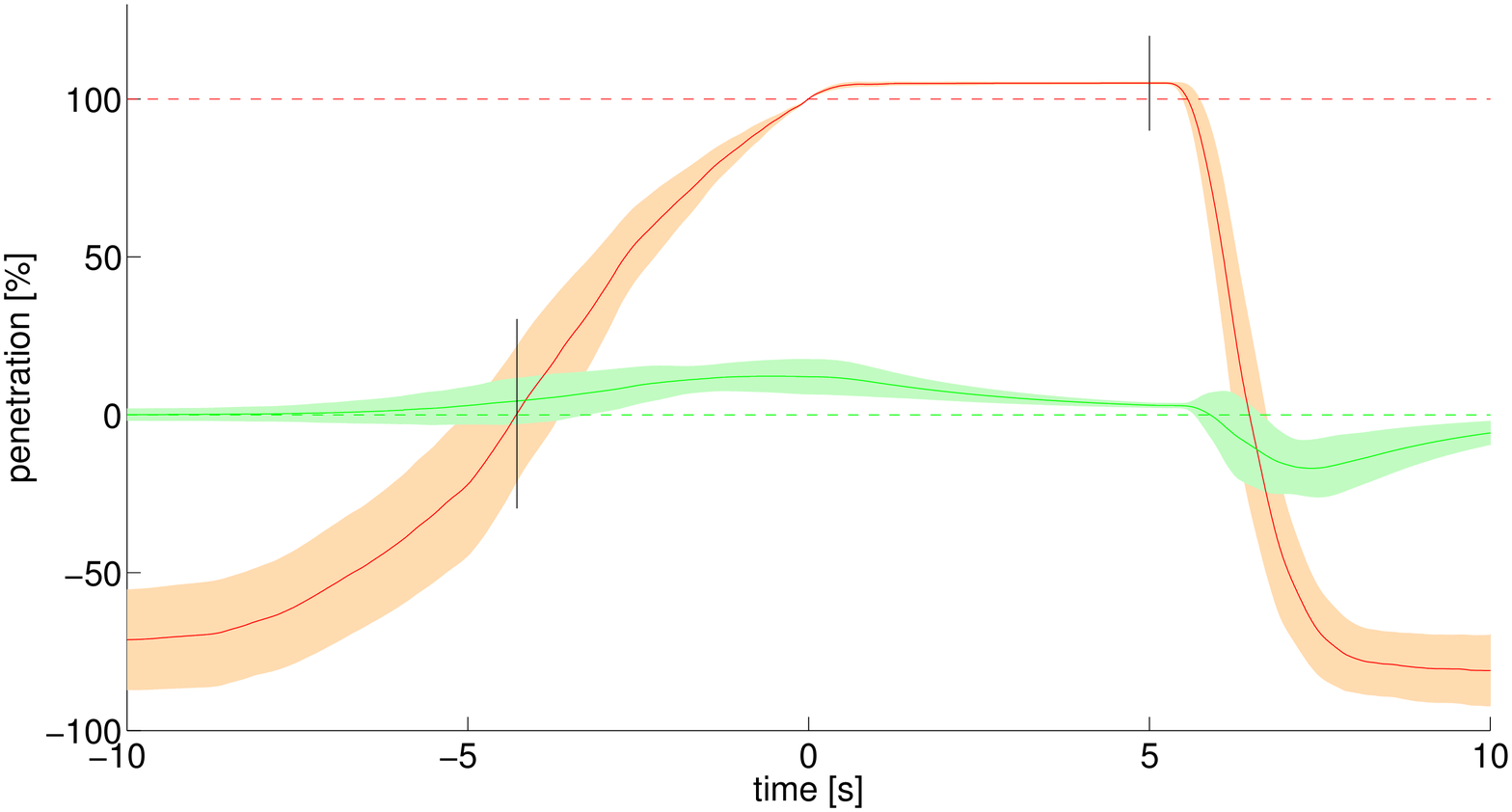}
		\label{nodelay:3}
	 }	 
	 
	 \subfloat[][Cutaneous feedback on the contralateral hand (CCF)]{
		\includegraphics[width=0.85\columnwidth]{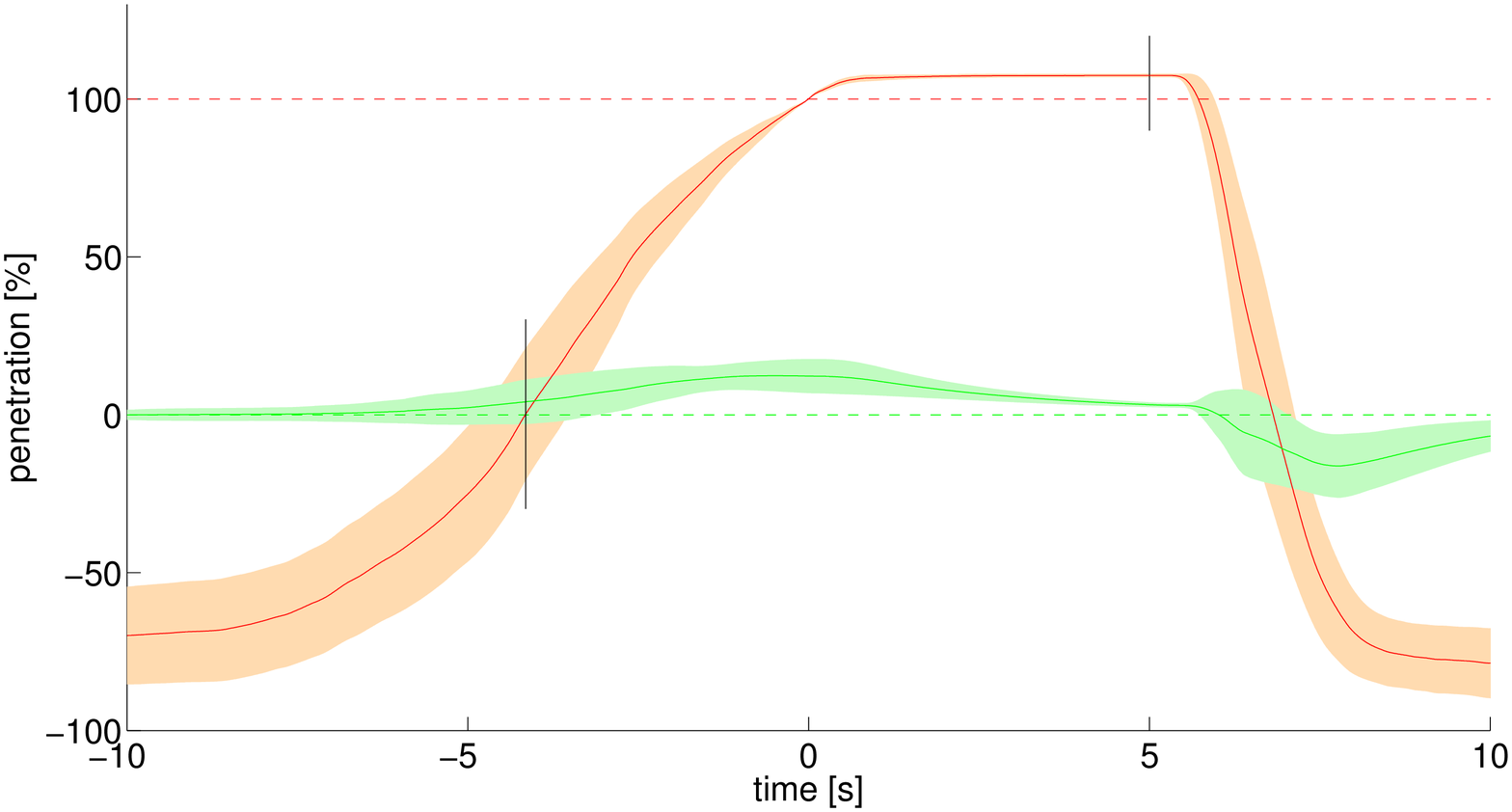}
		\label{nodelay:4}
	 }
\caption{Penetration of the needle (red patch) and position of tissue surface (green patch) versus time for experiment $ \#1 $. Average trajectories among subjects and their standard deviations are plotted. The position of the virtual fixture (dashed red line) and the initial position of tissue surface (dashed green line) are shown as well. The black lines represent the instants when the average trajectory enters the tissue (left line) and when the sound beep is played (right line).}
\label{nodelay}
\end{figure}

Three different experiments were implemented:

\begin{itemize}
\item \textit{experiment \#1}: twenty-four repetitions of the needle insertion task described above;
\item \textit{experiment \#2}: two additional repetitions of the needle insertion task, during which
      the position of the virtual fixture was changed suddenly and unexpectedly;
\item \textit{experiment \#3}: same as experiment \#1, but in presence of a time delay in the haptic loop.
\end{itemize}

The first experiment aimed at demonstrating that, on the one hand, there is no relevant degradation
of performance in the haptic interaction task (i.e., inserting the needle) when a normal force is
fed back to the user's fingertip holding the handle, using the cutaneous devices in substitution of
the feedback generated by a haptic device.  On the other hand, this experiment aimed at
demonstrating that using the cutaneous devices can lead to better performances with respect to other
forms of sensory substitution, such as visual feedback in substitution of force feedback, in which
the alternative feedback modality is different in nature from the one being substituted.  Moreover,
the experiment investigated if the fact that cutaneous force feedback provides a reliable form of
feedback is due \emph{only} to presenting a force to the user, or also to the fact that the feedback
information is applied to the fingertips which are responsible for handling the needle during the
experiments.

The second experiment aimed at showing that using the cutaneous devices prevents the handle (and so
the needle) from moving in unwanted directions in case of sudden and unpredictable changes of the
position of the virtual fixture.

The third experiment aimed at confirming the well known result that there are no instability
behaviours, not even in presence of delays, while using cutaneous force feedback devices.

\section{Experimental results}
\label{results}

\subsection{Experiment $\#1$: comparison of the feedback modalities}
\label{exp1}

Sixteen participants ($13$ males, $3$ females, age range $21$--$28$)
took part in the experiment, all of whom were right-handed. Eight of
them had previous experience with haptic interfaces.
None of the participants reported any deficiencies in the perception abilities
(including vision, hearing, touch and proprioception).
Each participant made $24$ repetitions of the needle insertion task,
with six randomized trials for each feedback mode:
\begin{itemize}
\item \textit{visual feedback} by the horizontal bar (task VF);
\item \textit{haptic feedback} (kinesthetic and cutaneous) by the haptic device (task HF);
\item \textit{cutaneous feedback} by the wearable devices, applied to the hand holding the handle (task CF);
\item \textit{cutaneous feedback} by the wearable devices, applied to the \textit{contralateral hand} (task CCF).
\end{itemize} 
The experiment lasted $ 9.13 $ minutes on average, including the two additional trials for experiment $ \#2 $, which followed the twelfth and the $24$th repetitions of experiment $ \#1 $ (see Sec. \ref{exp2} for details). A total of $ 26 $ tasks were performed by each subject,
$ 24 $ of which were included in the results of experiment $ \#1 $.

With the aim of comparing the different feedback modalities, the
position $z_{n}$ of the needle, steered by the operator's hand, was
recorded and the penetration into the virtual fixture $p =
\bar{z}_{vf} - z_{n}$ was calculated. The average penetration $\bar{p}$ and the maximum penetration
$\bar{p}_M$ were analyzed\footnote{Data resulting from different trials of the same task, performed by the same subject, were averaged before comparison with other tasks.}.
Such values provide a measure of accuracy (average penetration) and of overshoot (maximum penetration) in reaching the target depth. A null value in both metrics denotes the best performance, while a positive value indicates that the subject overrun the target. Both measures can be considered particularly relevant to the surgical task, as an excessive penetration of the needle can result in permanent damage of tissues.

Fig.~\ref{nodelay} shows the positions of the needle (red patch) and of the tissue surface (green patch) versus time. The time bases of different trials were synchronized at the time the needle first enters the fixture ($ t=0 $), while positions were divided by the depth of the virtual fixture, which varied randomly among trials, and are presented as percentage. Trajectories were averaged among subjects for each feedback modality, and average trajectories plus/minus standard deviations are shown. The position of the virtual fixture (dashed red line, $ 100$~percent) and the initial position of tissue surface (dashed green line, $ 0$~percent) are shown as well. The black lines represent the instants when the average trajectory enters the tissue (left line) and when the sound beep is played (right line).

\begin{figure}[t]
\begin{centering} 
\includegraphics[width=0.75\columnwidth]{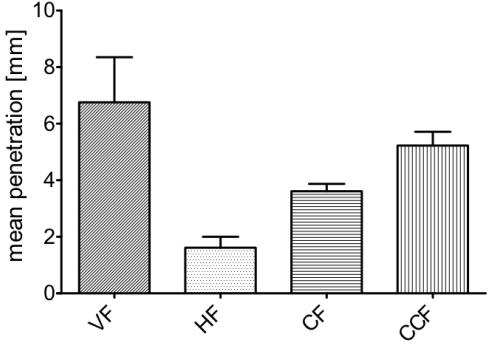}
\caption{Experiment $ \#1 $: average penetration beyond the virtual fixture (mean and SD), for the visual (VF), haptic (HF) and cutaneous feedback modes (CF, CCF). A null value of this metric indicates high accuracy in reaching the target depth.}
\label{means}
\end{centering}
\end{figure}

\begin{figure}[t]
\begin{centering}
\includegraphics[width=0.75\columnwidth]{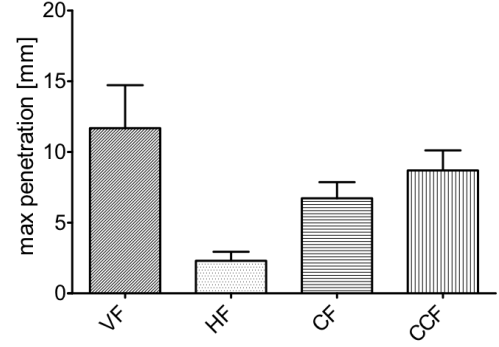}
\caption{Experiment $ \#1 $: maximum penetration beyond the virtual fixture (mean and SD), for the visual (VF), haptic (HF) and cutaneous feedback modes (CF, CCF). A null value of this metric indicates no overshoot in reaching the target depth.}
\label{maxs}
\end{centering}
\end{figure}

Fig.~\ref{means} and Fig.~\ref{maxs} show the average and maximum penetrations beyond the fixture for each feedback modality (means and standard deviations are plotted). All column data passed the D'Agostino-Pearson omnibus K2 normality test.
Comparison of the means among the feedback modalities was tested using one-way,
repeated measures analysis of variance (ANOVA).
The means of average penetration (Fig.~\ref{means}, $ F_{3, 45} = 106.5 $, $ P<0.0001 $)
and the means of maximum penetration (Fig.~\ref{maxs}, $ F_{3, 45} = 81.89 $, $ P<0.0001 $)
differed significantly among the feedback modalities.
Posthoc analyses (Bonferroni's multiple comparison test) revealed statistically significant difference between all pairs of columns,
both in terms of average penetration (Fig.~\ref{means}, $ P<0.001 $ for all pairs) and in terms of maximum penetration
of the needle (Fig.~\ref{maxs}, $ P<0.05 $ for CF vs CCF, and $ P<0.001 $ for all other pairs).
Results indicate that the proposed sensory subtraction modality (CF) yields an intermediate performance between haptic feedback (HF, best performance) and visual feedback (VF, worst case), in terms of both average and maximum penetration beyond the virtual fixture.
These results demonstrate also that the cutaneous devices provide a more reliable form of feedback if applied to the fingertips which are responsible for holding the end-effector (CF) with respect to contralateral hand stimulation (CCF), suggesting that the localization of cutaneous feedback is crucial in this setting.
Nonetheless, cutaneous feedback is more efficacious than visual feedback (VF) even when it is applied to the contralateral hand (CCF), indicating that not only the localization but also the nature of the sensation provided is relevant to task performance.

\begin{figure}[t]
\begin{centering}
\includegraphics[width=0.75\columnwidth]{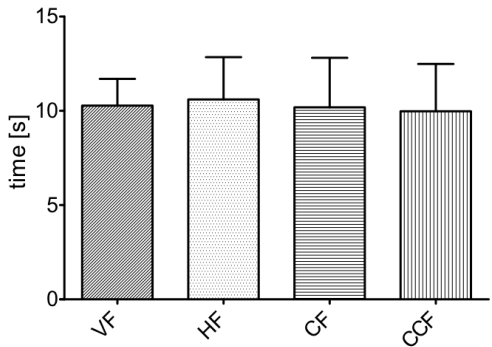}
\caption{Experiment $ \#1 $: time elapsed (mean and SD) between the first contact with the tissue and the sound played after $5$~s of continuous contact with the virtual fixture, for the visual (VF), haptic (HF) and cutaneous feedback modes (CF, CCF).}
\label{tempi}
\end{centering}
\end{figure}

Fig.~\ref{tempi} shows the average time elapsed between the instant
the needle penetrates the tissue and the instant it reaches $5$~s of
continuous contact with the virtual fixture.
Column data failed to pass the normality test, so the Friedman non-parametric test was used to analyze variance. Results indicate that there is no statistically significant difference between
the feedback modalities in this metric ($ P>0.1 $).
We may read this result by saying that the subjects became equally
confident with all the feedback modalities proposed.

\subsection{Experiment $\#2$: dynamic virtual fixture}
\label{exp2}

This experiment evaluated the effect on needle position of a sudden and unpredictable change of the position of the virtual fixture, in the presence of the four feedback modalities described before (visual, haptic and the two cutaneous).
In this new test, the needle insertion task was the same as that described in Sec.~\ref{exp1}. However, after $5$~s of continuous contact, the depth of the virtual fixture was increased unexpectedly, so the virtual environment suddenly fed back no guiding force to the user.
At the same time, the sound beep was produced as in the other repetitions of the needle insertion task, signaling the subject to extract the needle.
The two circumstances provided conflicting information to the user. In fact, the user was initially instructed to keep contact with the fixture, so at the one hand the sudden change in the guiding force suggested to increase needle depth. On the other hand, the sound signaled to extract the needle.

The test was performed during two additional trials of experiment $\#1$. To ensure the surprise effect, each of the subjects who took part in the experiment $\#1$ performed only two additional trials (using two different feedback modalities). A~total of $ 32 $ trials were recorded for experiment $ \#2 $: $8$ trials per each feedback modality, performed by eight different subjects. 
The first additional trial was run after the $12$th trial of experiment $\#1$, the second after the $24$th. %
No information was provided to the subjects about the additional trials, which followed immediately the previous ones. A $ 30$~s rest was given to all subjects after the first additional trial, before continuing with the second part of experiment $\#1$.
Subjects did not know that the position of the virtual fixture was going to change and that they were performing a different task with respect to the others.

\begin{figure}[t]
\begin{centering}
\includegraphics[width=0.75\columnwidth]{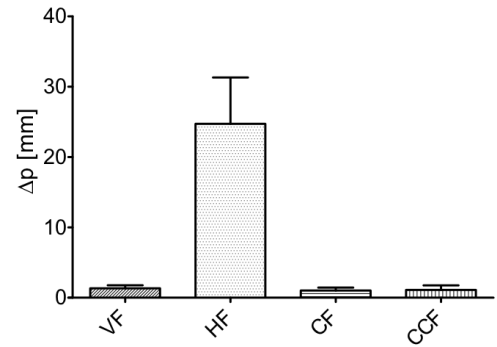}
\caption{Experiment $ \#2 $: difference (mean and SD) between the maximum penetration, after the movement of the virtual fixture, and the average penetration registered before (during continuous contact), for the visual (VF), haptic (HF) and cutaneous feedback modes (CF, CCF).}
\label{peakdiff}
\end{centering}
\end{figure}

Fig.~\ref{peakdiff} shows the differences $ \Delta p $ between the maximum penetration registered after the perturbation and the average penetration observed in the $5$~s before (continuous contact). 
All column data passed the D'Agostino-Pearson omnibus K2 normality test.
Comparison of the means among the feedback modalities was tested using one-way ANOVA (no repeated measures).
The means differed significantly among the feedback modalities ($ F_{3, 28} = 100.3 $, $ P<0.0001 $). Posthoc analyses (Bonferroni's multiple comparison test) revealed statistically significant difference between haptic feedback (HF) and each alternate modality (VF, CF, CCF, $ P<0.001 $).
Results indicate that the presence of kinesthetic feedback may induce significantly greater unwanted motions of the needle with respect to the three non-kinesthetic feedback modes used in the experiments (the visual and the two cutaneous-only modalities). In fact, when the fixture moves in haptic mode (HF), the subject's arm is counteracting an external force which suddenly drops.

\begin{figure}[]
\begin{centering}
 \subfloat[][Visual feedback (VF)]{
  \label{vfmove:1}
\includegraphics[width=0.85\columnwidth]{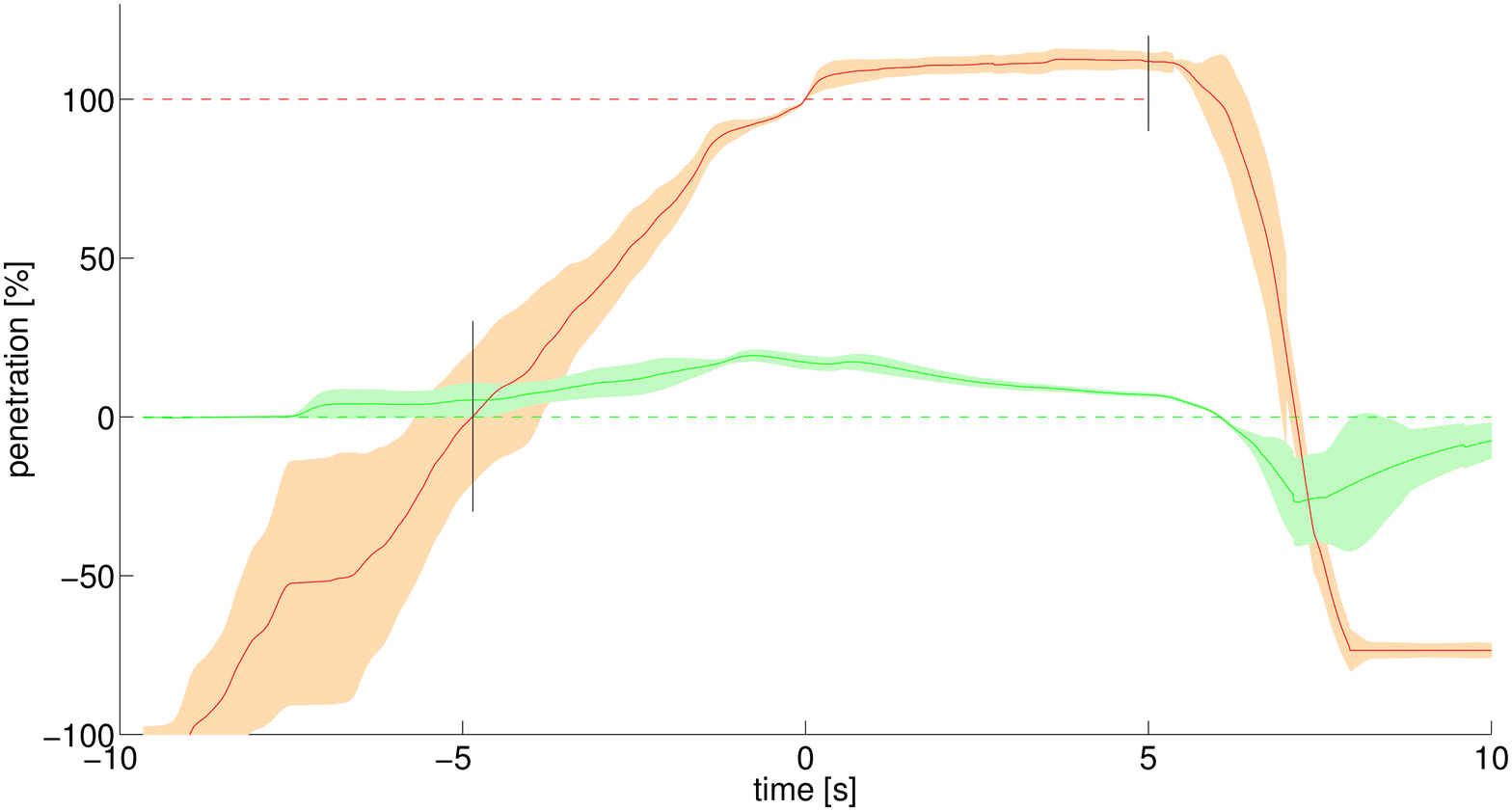}
}

 \subfloat[][Complete haptic feedback (HF)]{
  \label{vfmove:2}
\includegraphics[width=0.85\columnwidth]{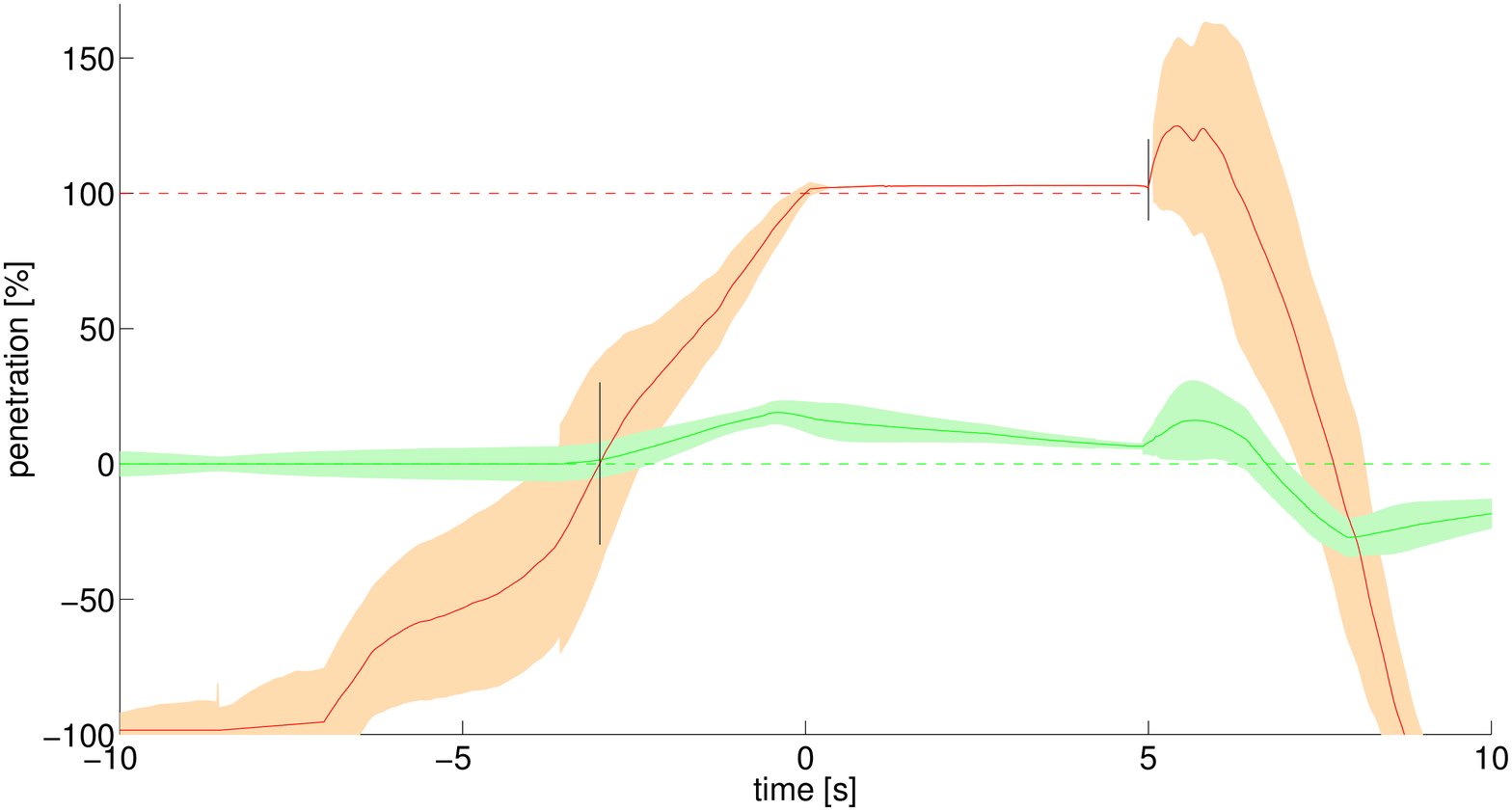}
}

 \subfloat[][Cutaneous feedback on the hand holding the handle (CF)]{
  \label{vfmove:3}
\includegraphics[width=0.85\columnwidth]{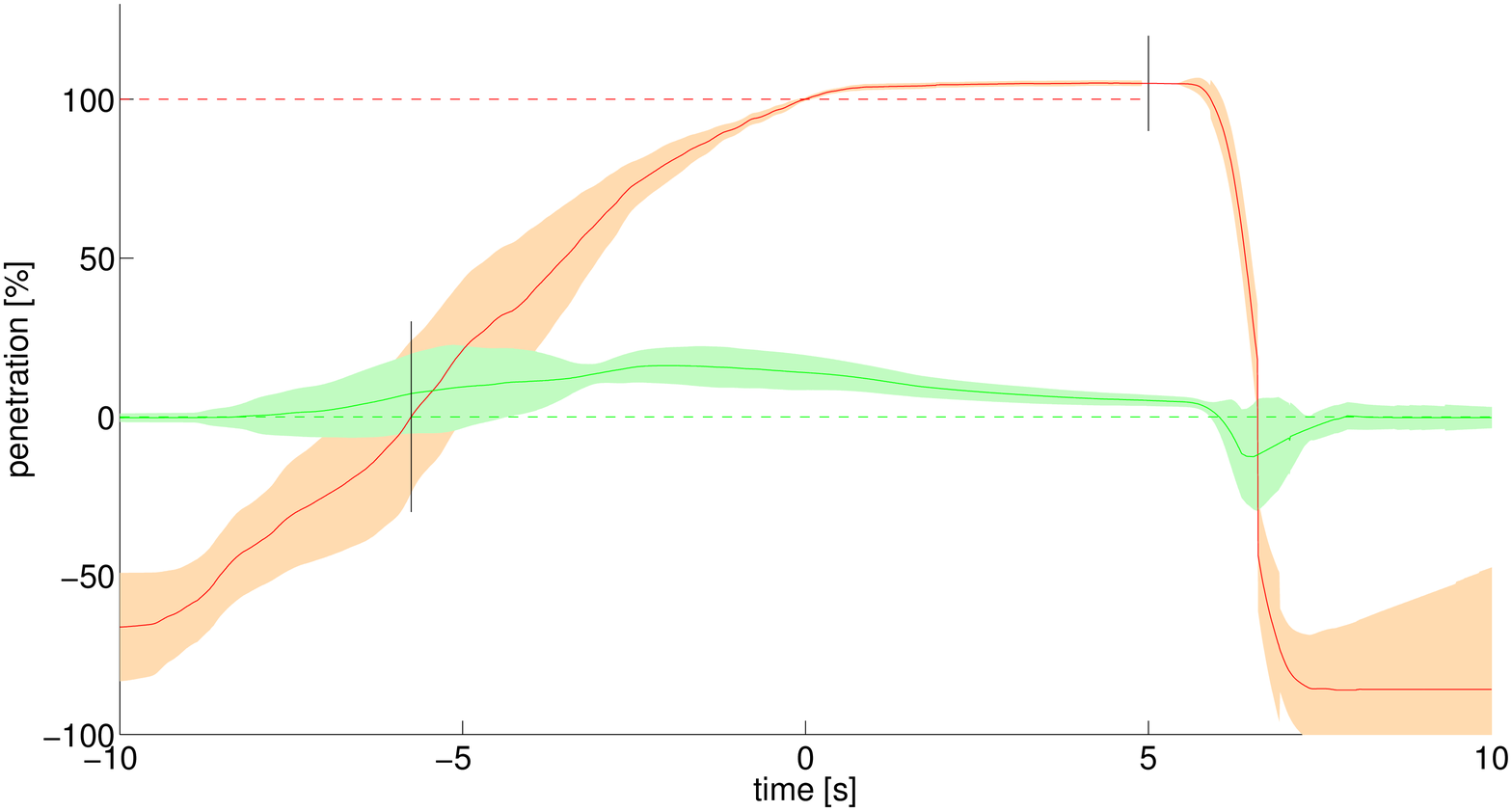}
}

 \subfloat[][Cutaneous feedback on the contralateral hand (CCF)]{
  \label{vfmove:4}
\includegraphics[width=0.85\columnwidth]{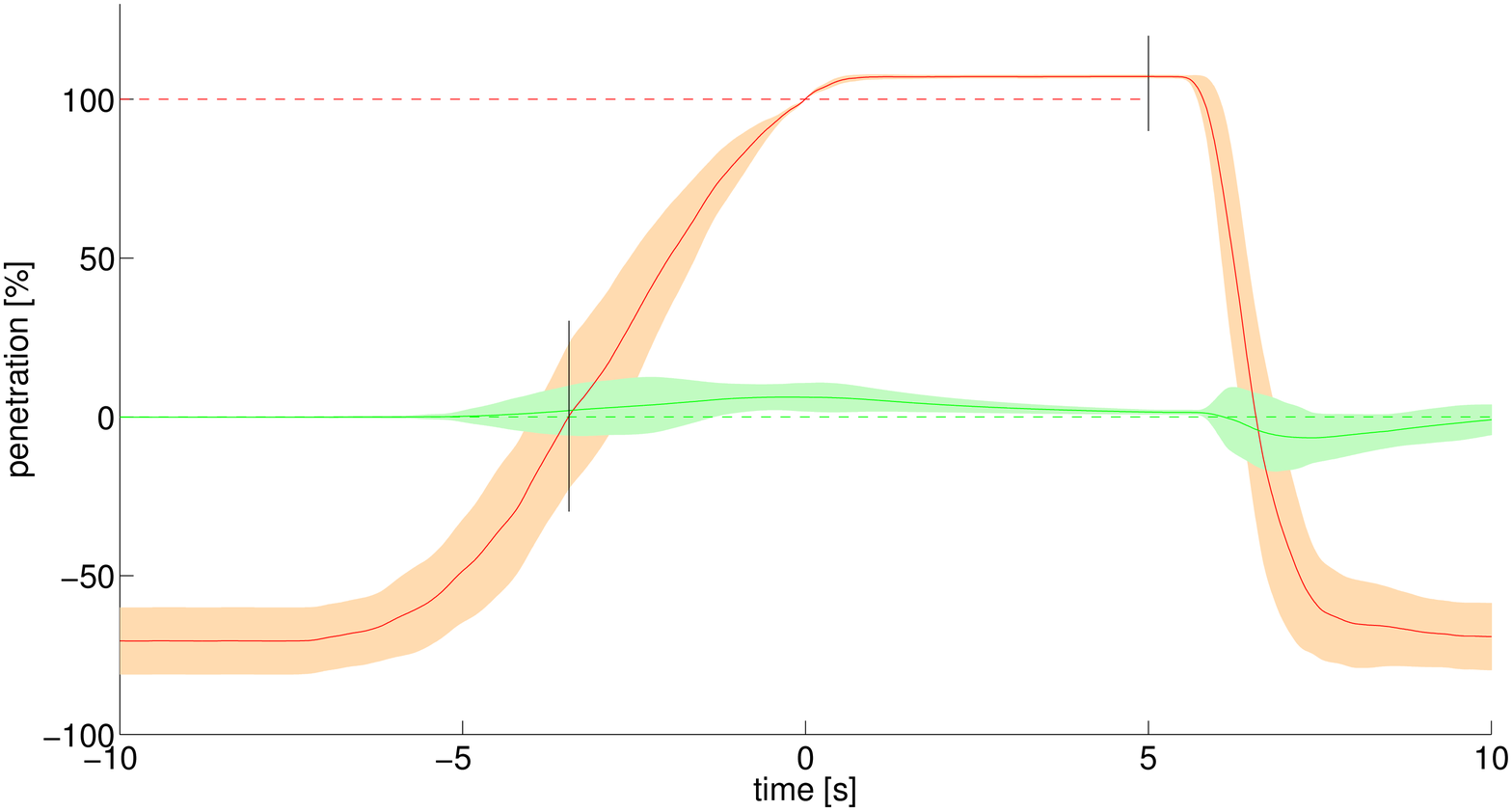}
}
\caption{Penetration of the needle (red patch) and position of tissue surface (green patch) versus time for experiment $ \#2 $, with the virtual fixture suddenly removed after $ 5$~s of continuous contact. Average trajectories among subjects and their standard deviations are plotted. The position of the virtual fixture (dashed red line) and the initial position of tissue surface (dashed green line) are shown as well. The black lines represent the instants when the average trajectory enters the tissue (left line) and when the virtual fixture is removed and the sound beep is played (right line).}
\label{vfmove}
\end{centering}
\end{figure}

Fig.~\ref{vfmove} shows the positions of the needle (red patch) and of the tissue surface (green patch) versus time for all the groups of experiment $ \#2 $. Data were synchronized, normalized and averaged among subjects as for the charts of Fig.~\ref{nodelay}.

\subsection{Experiment $\#3$: stability with time delay}
\label{exp3}

As other sensory substitution techniques, the main advantage of
the proposed cutaneous--feedback sensory subtraction is that
it makes the haptic loop intrinsically stable. No instability
behaviors occur, even in presence of large delays.

To support this hypothesis, a new set of experiments was implemented,
in which the same protocol used in the experiment described in
Sec.~\ref{exp1} was used for the needle insertion task, including the types of feedback
employed and number of repetitions ($ 24 $) per subject, but here a delay of
$50$~ms was introduced in the haptic loop between the virtual
environment and either the haptic handle, the cutaneous devices or the
visual rendering of force. Recent literature denotes the relevance of delays in teleoperated surgical tasks \cite{2011MussaIvaldi}. It is worth noting that instability of
haptic feedback in the presence of time delays can be fixed with a
wave variable transformation
\cite{niemeyer1991,ye2009,lam2008}. Nonetheless, to emphasize the intrinsic
stability of cutaneous feedback, this method was not used in the trials.

Ten participants ($8$ males, $2$ females, age range $20$--$26$) took
part in the experiment, all of whom were right-handed and five of
whom had previous experience with haptic interfaces. None of the
participants reported any deficiencies in the perception abilities
(as defined before).
The experiment lasted $ 8.39 $ minutes on average.

\begin{figure}[]
\centering
 \subfloat[][Visual feedback (VF)]{
  \label{delay:1}
\includegraphics[width=0.85\columnwidth]{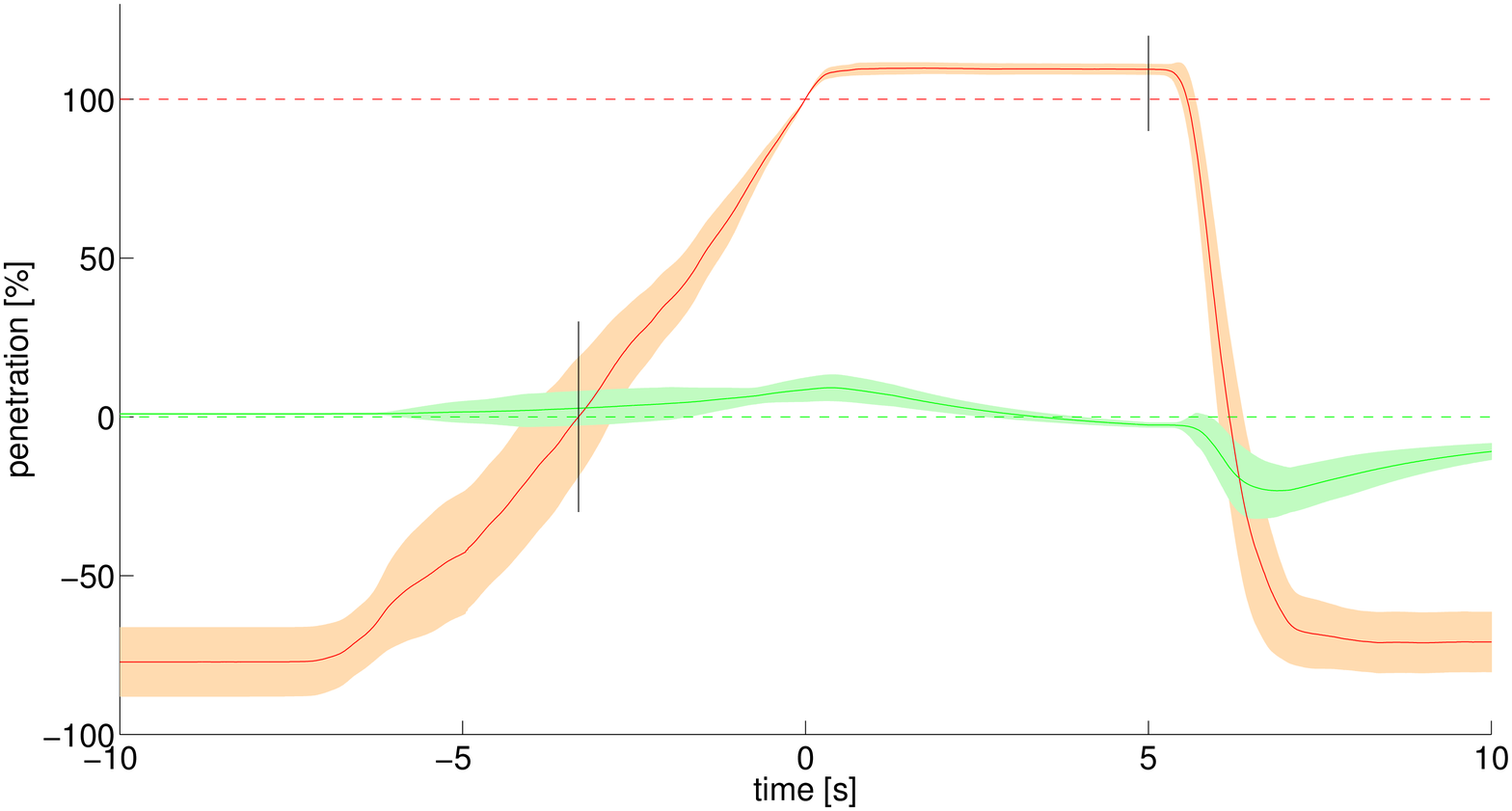}
}

 \subfloat[][Complete haptic feedback (HF)]{
  \label{delay:2}
\includegraphics[width=0.85\columnwidth]{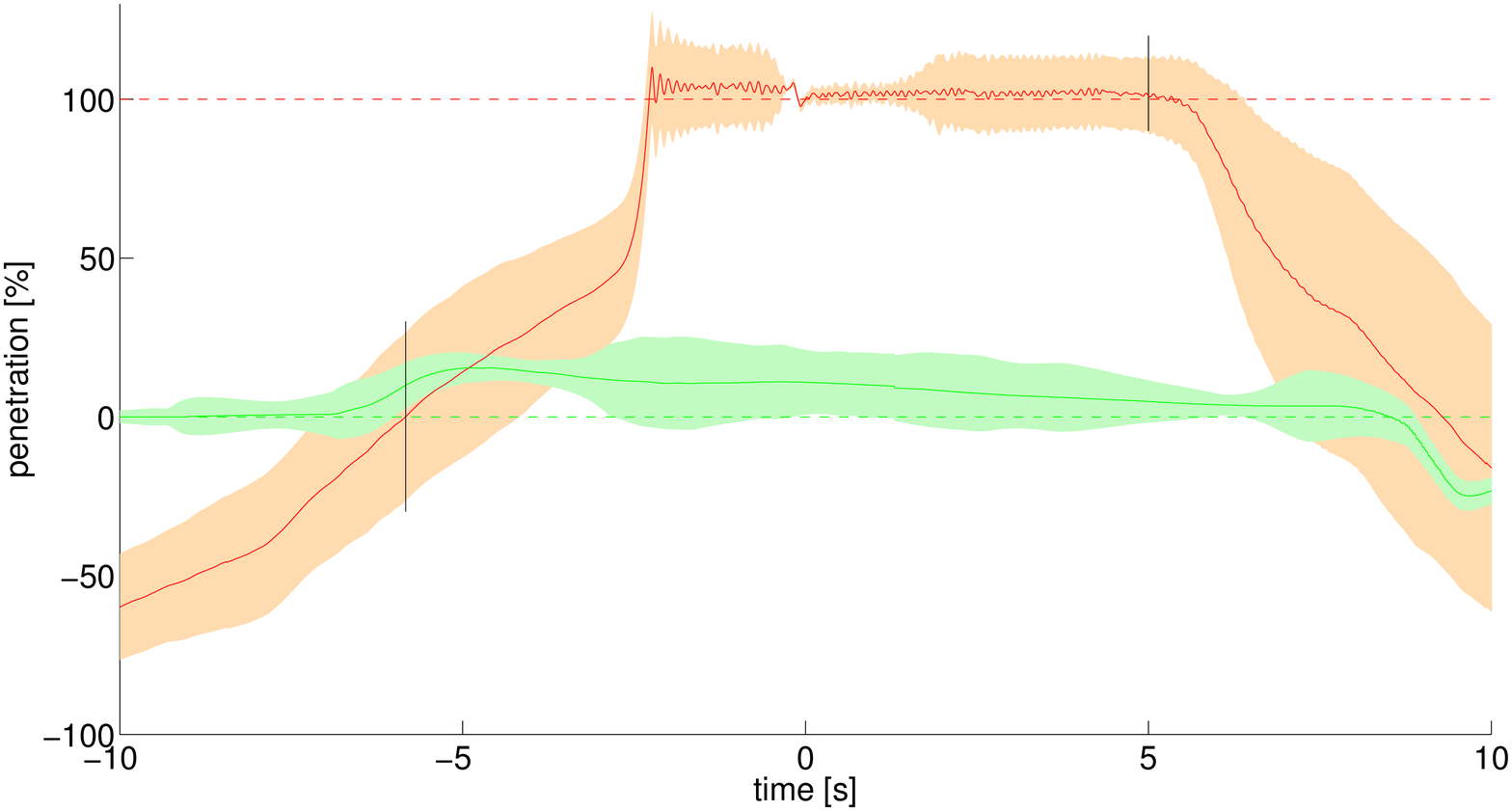}
}

 \subfloat[][Cutaneous feedback on the hand holding the handle (CF)]{
  \label{delay:3}
\includegraphics[width=0.85\columnwidth]{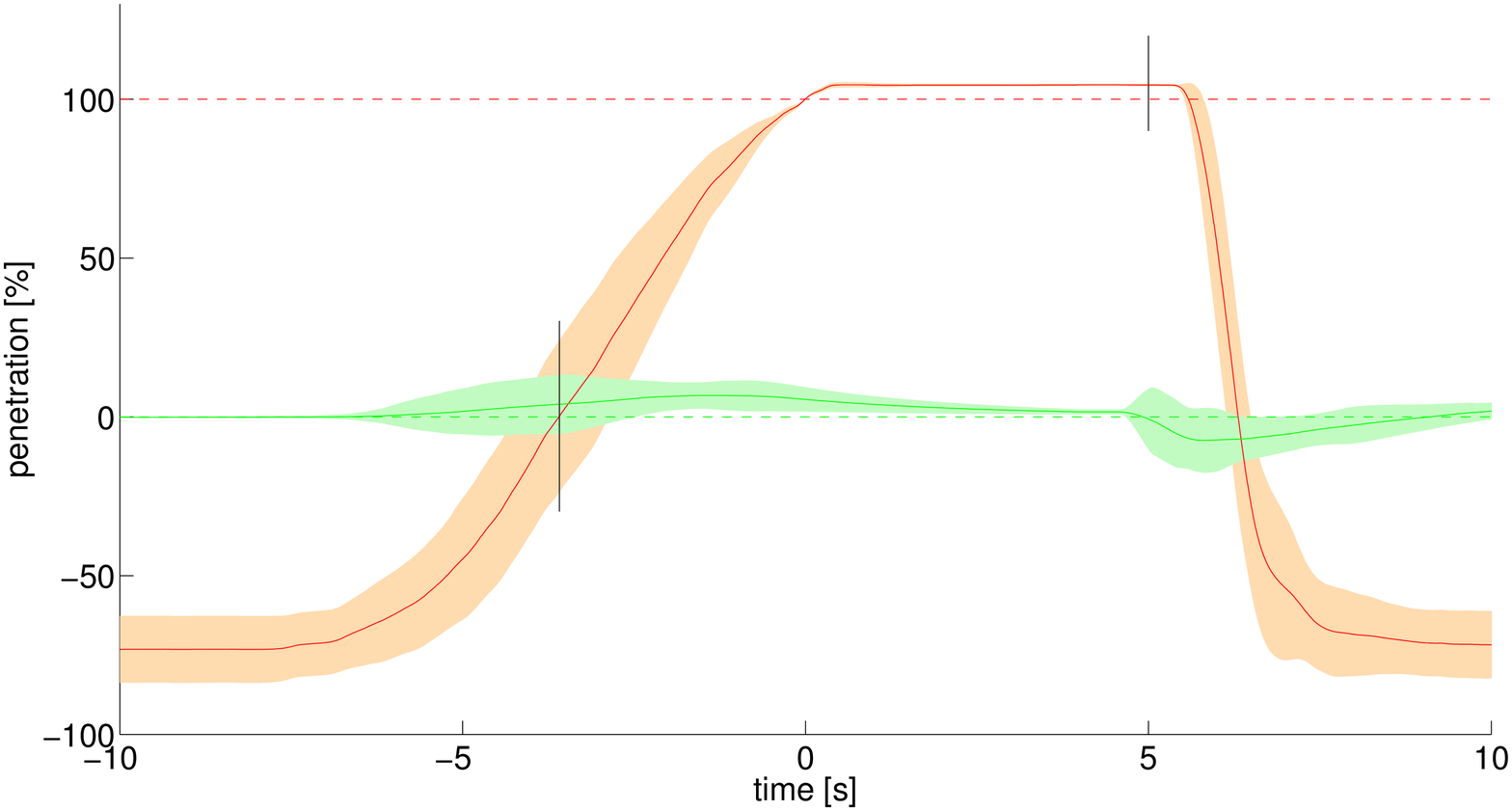}
}

 \subfloat[][Cutaneous feedback on the contralateral hand (CCF)]{
  \label{delay:4}
\includegraphics[width=0.85\columnwidth]{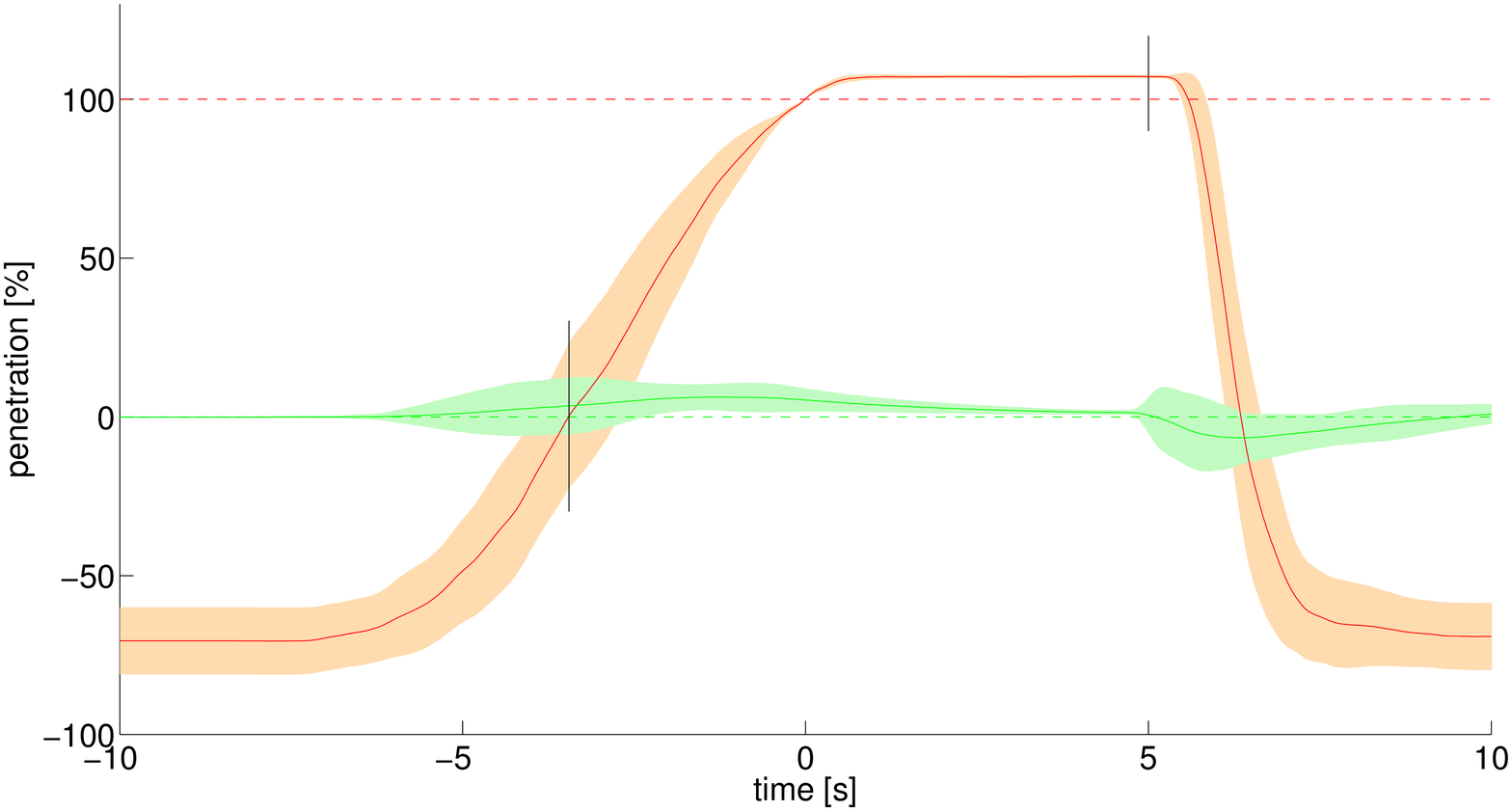}
}
\caption{Penetration of the needle (red patch) and position of tissue surface (green patch) versus time for experiment $ \#3 $, with a $50$~ms network delay in the haptic loop. Average trajectories among subjects and their standard deviations are plotted. The position of the virtual fixture (dashed red line) and the initial position of tissue surface (dashed green line) are shown as well. The black lines represent the instants when the average trajectory enters the tissue (left line) and when the sound beep is played (right line).}
\label{delay}
\end{figure}

Fig.~\ref{delay} shows the positions of the needle (red patch) and of the tissue surface (green patch) versus time for experiment $ \#3 $. Data were synchronized, normalized and averaged among subjects as for Fig.~\ref{nodelay}. 
By comparing the charts with those in Fig.~\ref{nodelay}, we can notice
that the instability occurred only with haptic feedback,
i.e., only in the presence of kinesthetic feedback. Significant
oscillations of the needle are likely to bring not only a greater
penetration of the needle in the virtual fixture, but also a longer
task completion time\footnote{A short movie of an experimental run showing the instability issue can be downloaded at \texttt{http://goo.gl/9vDqC}}.

\begin{figure}[!t]
\centering 
\includegraphics[width=0.75\columnwidth]{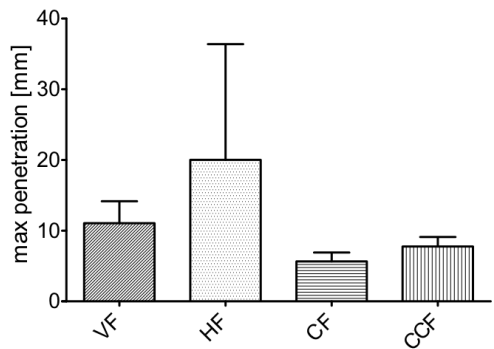}
\caption{Experiment $ \#3 $: maximum penetration beyond the virtual fixture (mean and SD), for the the visual (VF), haptic (HF) and cutaneous feedback modes (CF, CCF), with a $50$~ms network delay in the loop.}
\label{dmaxs}
\end{figure}

Fig.~\ref{dmaxs} shows the maximum penetration beyond the fixture in the presence of the delay.
Haptic feedback group data (HF) failed to pass the normality test, so the Friedman non-parametric test was used to analyze variance. The test indicated statistically significant difference between the feedback modalities ($ P<0.0001 $).
Posthoc analyses (Dunn's multiple comparison test) revealed statistically significant difference between haptic feedback (HF) and both cutaneous modalities (CF, $ P<0.001 $; CCF, $ P<0.05 $) and between cutaneous feedback (CF) and visual feedback (VF, $ P<0.001 $).
Results indicate that the subjects, while receiving the complete haptic feedback in the presence of a time delay, reached a significantly greater peak penetration in the virtual fixture with respect to that obtained while receiving feedback from the wearable cutaneous devices, regardless the localization of cutaneous feedback.
The same result was obtained when the subjects received visual feedback of force instead of cutaneous feedback on the fingers which are responsible for handling the needle.

\begin{figure}[!t]
\centering 
\includegraphics[width=0.75\columnwidth]{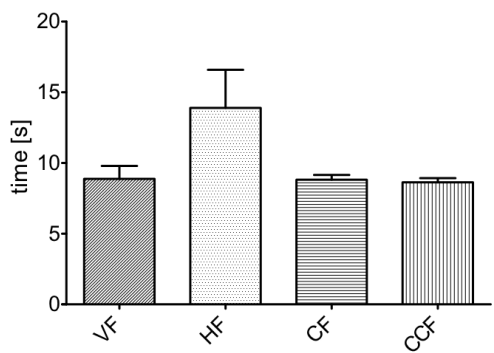}
\caption{Experiment $ \#3 $: time elapsed between the first contact with the tissue and the sound
  beep, for the visual (VF), haptic (HF) and cutaneous feedback modes (CF, CCF), with a $50$~ms
  network delay in the haptic loop.}
\label{dtempi}
\end{figure}

Fig.~\ref{dtempi} shows, for each feedback modality and in the presence of the time delay, the mean time elapsed between the first penetration in the tissue and the instant the needle reaches $5$~s of stable contact with the virtual fixture.
Haptic feedback and cutaneous feedback group data failed to pass the normality test, so the Friedman non-parametric test was used to analyze variance. The test indicated statistically significant difference between the feedback modalities ($ P<0.0001 $).
Posthoc analyses (Dunn's multiple comparison test) revealed statistically significant difference between haptic feedback (HF) and all other feedback modalities (VF, $ P<0.01 $; CF, $ P<0.05 $; CCF, $ P<0.001 $).
Results indicate that the time needed to accomplish the task was significantly greater while receiving the kinesthetic feedback with respect to the other non-kinesthetic feedback modalities.
Such a difference had not been observed in the absence of time delays (Fig.~\ref{tempi}), and must be related to instability.

\subsection{Discussion}
\label{discussion}
The first experiment evaluated the effectiveness of the sensory
subtraction technique proposed in the paper.  The results of this
experiment indicate that the subjects, while receiving visual feedback (VF)
in substitution of force feedback, reached a significantly
greater average and maximum penetration in the virtual fixture (worst performance) in comparison with that obtained while receiving either complete haptic (HF) or cutaneous-only
feedback (CF and CCF). The last two modalities provided intermediate performance between visual and haptic feedback. No difference between groups was observed in terms of task completion time.

As expected, haptic feedback outperformed all the other feedback modes. The cutaneous-only modality
proved itself to be a more intuitive form of feedback than other sensory substitution techniques,
regardless the localization of the cutaneous devices (either on the hand performing the task or on
the contralateral hand).
When the cutaneous force feedback was applied to the contralateral hand (i.e., the one not involved
in controlling the motion of the input device), performance was worst in terms of penetration of the
virtual fixture with respect to the case when the cutaneous force feedback was applied to the acting
hand.  A possible mechanistic interpretation could be that the cutaneous feedback applied to the
contralateral hand needs time for transcallosal transmission to reach the hemisphere controlling the
operating hand.  In fact, the feedback reaches the hemisphere of the brain not involved in the motor
control of the hand moving the input device, and for this reason requires more time to be
transformed in motor action \cite{kandel2000principles}.

It is worth underlying that larger penetration into the virtual fixture corresponds to a higher
force fed back by the virtual environment, applied by either the haptic device, the cutaneous
actuators or displayed using the horizontal bar for sensory substitution with visual modality.  Also
note that the larger penetration observed when cutaneous force feedback was used may be partly due
to the delay of the cutaneous actuators employed in the tests, which can be quantified in $ \sim45$~ms.


These results suggest that the novel feedback modality can be successfully used in substitution of
traditional haptic feedback, with a minor decay of performance with respect to visual sensory
substitution techniques.  Not only the type of feedback, cutaneous rather than visual, but also the
place where it is applied is important. The best performance is obtained when the cutaneous devices
are worn on the hand involved in the task, i.e., when they provide the user with a subset of the
stimuli produced by the input device in complete haptic mode.  This result can be explained by
considering that the area of application of the force and the particular design of the cutaneous
devices proposed provide the user with a direct and intuitive measure of the contact force being
substituted, thus producing a more natural interaction with the device.

One interesting result observed with sensory subtraction is that, during the first experiment, performance degraded when kinesthetic information was removed. One possible explanation is that the external force subtracted played a role in arm dynamics during the execution of the needle insertion task. In particular, the virtual force helped the subject in stopping hand motion when the virtual fixture was reached, which is the main reason for using virtual fixtures indeed. Conversely, in the proposed touch-only modality, no physical aid is provided to the user to accomplish the task, so arm motion derives entirely from motor control. The resulting benefit is that unwanted motions can be drastically reduced in critical situations.

On the other hand, without adequate sense of touch, achieving normal and top performance in tasks that require high levels of dexterity is extremely difficult, if not impossible \cite{2006DeLaTorre}. Moreover, even simple touch information can be effective both in virtual and in real environments.
For example, major gains in body posture control in real environments can be obtained from minimal touch information applied to a fingertip \cite{1998Jeka}. This may explain why touch-only tasks were  better executed than the substituted visual tasks.

One major advantage of sensory subtraction is that, despite the fact
that the interaction is closer to haptic rendering, no unwanted
movements are likely to be produced during the execution of guided
tasks. This achievement, that is corroborated by the results of
experiment $\#2$, is particularly crucial in critical applications
such as robot-aided surgery, in which unwanted movements of the
surgeon's hand induced by force feedback may produce serious damages
to the patient. The absence of unwanted movements, even in the case of
sudden and unpredictable changes of the position of the virtual
fixture, can be explained by considering that kinesthetic feedback was
completely eliminated in the cutaneous feedback modality, so the user
could maintain a stable contact with the virtual fixture without
exerting an active force on the handle.

The last experiment showed that, in the presence of a transmission delay,
complete haptic feedback can bring the haptic loop near to
instability, as significant oscillations of needle position occurred,
whereas cutaneous (and visual) feedback allows a stable contact with the
virtual fixture surface. The occurrence of instability with a relatively small time delay
may be due to the particular setting of the experimental device used in the
experiments. However, the fact that kinesthesia can bring instability
in haptic teleoperation in the presence of time delays is well-know in the
literature on haptics as discussed in the introductory section.

Another drawback of using complete haptic feedback in presence
of transmission delays is the longer time needed to complete the
task. Statistical analysis on task completion times showed that, in case of no delay,
there are no significant differences between the four different feedback modalities,
while in the presence of a network delay,
task completion time using haptic feedback can be significantly
greater than that obtained using cutaneous-only feedback.

\section{Conclusion and future work}
\label{conclusions}

We showed that cutaneous force feedback applied to the thumb and index finger pads during the
manipulation of a handle in teleoperation tasks can be effectively used to substitute complete
cutaneous and kinesthetic feedback in haptics. The main advantage of using cutaneous force feedback
displays is that the stability of the haptic loop is intrinsically guaranteed. This can be very
convenient for critical applications like robotic surgery. Note also that actuation for cutaneous
displays usually requires less power and it is less bulky than that required to provide haptic
feedback, with a direct effect on simplifying mechanical design and reducing costs.  

The main drawback of the proposed approach is that, like for other sensory substitution techniques,
the realism of the interaction is weaker when compared to complete haptic feedback but, differently
from other substitution techniques, the proposed one has the advantage of being perceived exactly
where it is expected and provides the operator with a direct and co-located perception of the
contact force even if it is only cutaneous and not kinesthetic. This is a possible explanation of
the better performances of the proposed sensory subtraction technique.


Although the mechanical design leads to simple light and portable cutaneous devices, work is in
progress to improve their level of wearability thus reducing the impact of using such devices.
%
%
%
%
Work is in progress to design new cutaneous displays with better dynamic performances, in order to
design and conduct additional psychophysical experiments to assess other relevant parameters, like
for instance the just noticeable difference (JND) for mechanical properties \cite{2005_Hirche}.
Another important aspect of future research is to evaluate the possibility of presenting to the user
not only cutaneous cues but also the kinesthetic feedback with a scaled intensity.  Setting the
scaling factor will be an interesting aspect of this research.  Also the combination of the
cutaneous-only paradigm with other modalities, like auditory feedback, is worthy being investigated.
Finally, while in this study we did not consider the possibility of applying any vibratory signal to
the cutaneous display since this was not compatible with the sensory subtraction idea, work is in
progress to compare the cutaneous force feedback, as driven by the sensory subtraction technique, to
other sensory substitution techniques using vibrotactile signals.

\section*{Acknowledgments}
This work was partly supported by the European Commission with the
      Collaborative Project grant FP7- ICT-2009-6-270460 ``ACTIVE: Active Constraints Technologies
      for Ill- defined or Volatile Environments''. The authors are grateful to Rudy Manganelli and
      Francesco Chinello from University of Siena for their help in realizing the experiments.

\bibliographystyle{IEEEtran}
\bibliography{biblio}

\vspace*{-1cm}

\begin{biography}[{\includegraphics[width=1in,height=1.25in,clip,keepaspectratio]{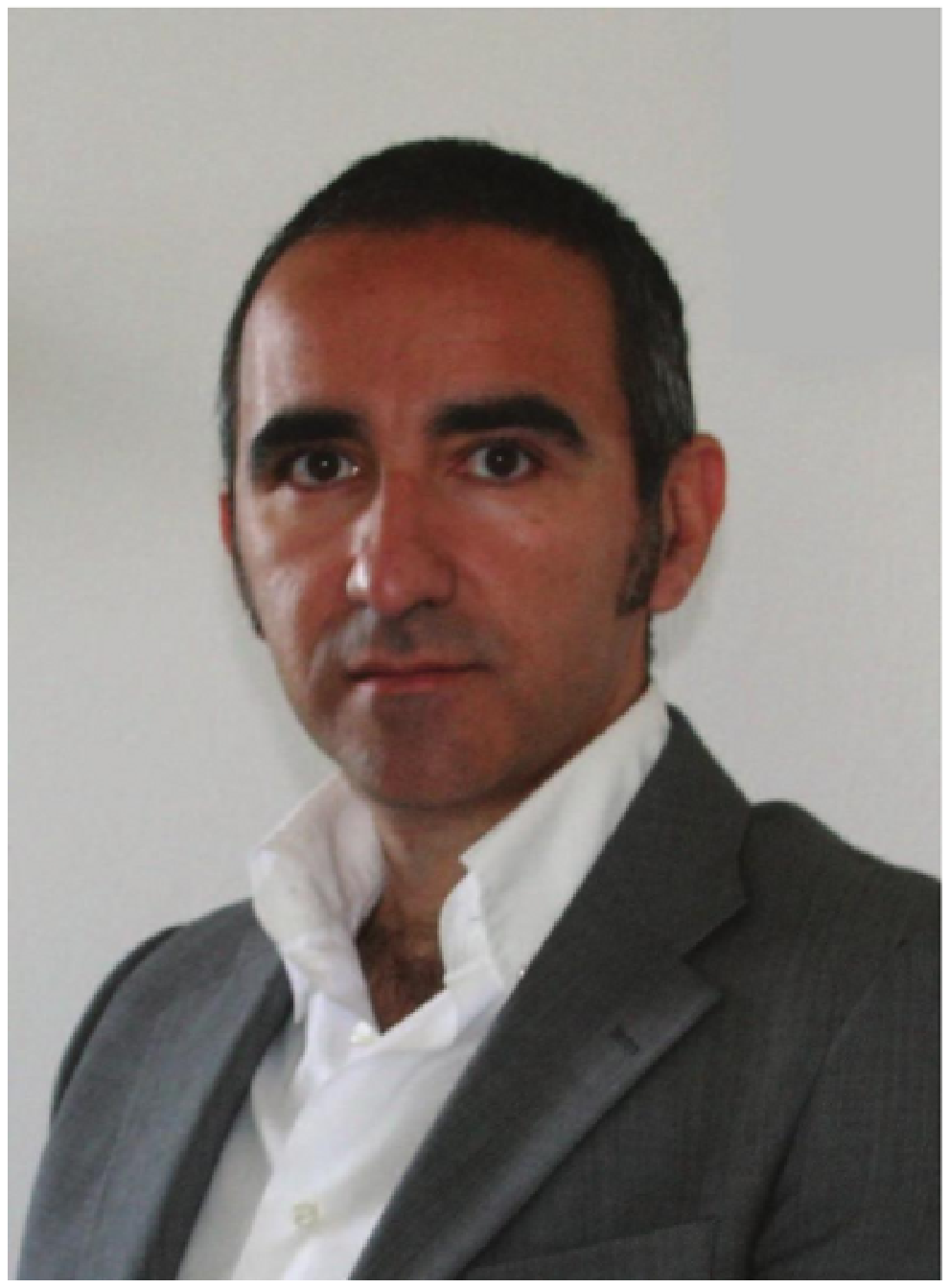}}]{Domenico
    Prattichizzo} (S'93 - M'95) received the M.S. degree in Electronics Engineering and the
  Ph.D. degree in Robotics and Automation from the University of Pisa in 1991 and 1995,
  respectively.  Since 2002 Associate Professor of Robotics at the University of Siena. Since 2009
  Scientific Consultant at Istituto Italiano di Tecnologia, Genova Italy. In 1994, Visiting
  Scientist at the MIT AI Lab.  Co-author of the Grasping chapter of Handbook of Robotics Springer,
  2008, awarded with two PROSE Awards presented by the American Association of Publishers.  Since
  2007 Associate Editor in Chief of the IEEE Trans. on Haptics. From 2003 to 2007, Associate Editor
  of the IEEE Trans on Robotics and IEEE Trans. on Control Systems Technologies. Vice-chair for
  Special Issues of the IEEE Technical Committee on Haptics (2006-2010). Chair of the Italian
  Chapter of the IEEE RAS (2006-2010), awarded with the IEEE 2009 Chapter of the Year Award.
  Co-editor of two books by STAR, Springer Tracks in Advanced Robotics, Springer (2003,
  2005). Research interests are in haptics, grasping, visual servoing, mobile robotics and geometric
  control. Author of more than 200 papers in these fields.
\end{biography}

\vspace*{-1cm}

\begin{biography}[{\includegraphics[width=1in,height=1.25in,clip,keepaspectratio]{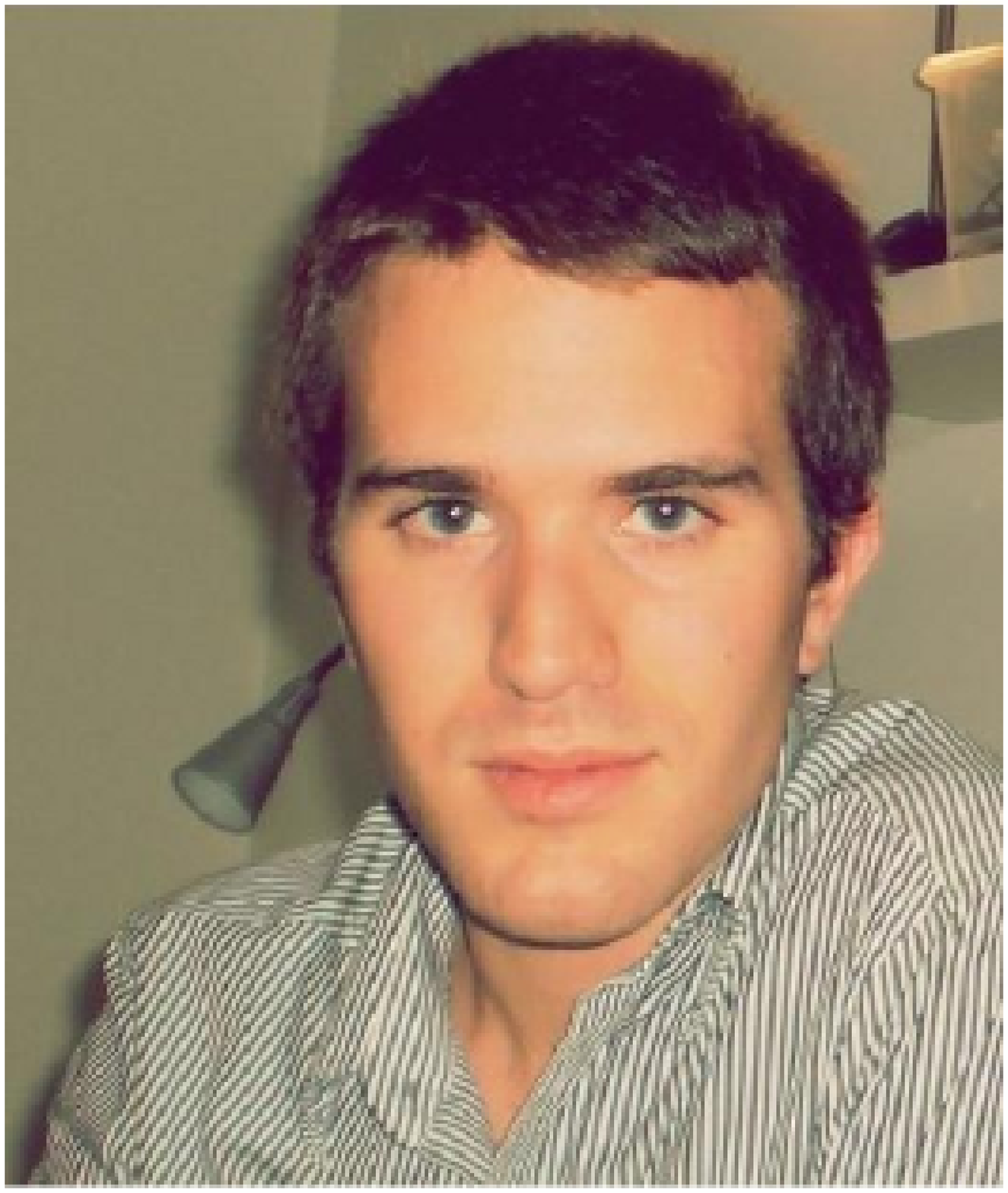}}]{Claudio
    Pacchierotti} (S'12) received the M.S. degree \emph{con lode} in computer engineering in 2011
  from the University of Siena, Italy. He was an exchange student at the Karlstad University, Sweden
  in 2010. He is currently a Ph.D. student at the Department of Information Engineering of the
  University of Siena and at the Department of Advanced Robotics of the Italian Institute of
  Technology.  His research interests include robotics and haptics, focusing on cutaneous force
  feedback techniques and human-robot interfaces for medical applications.
\end{biography}

\vspace*{-1cm}

\begin{biography}[{\includegraphics[width=1in,height=1.25in,clip,keepaspectratio]{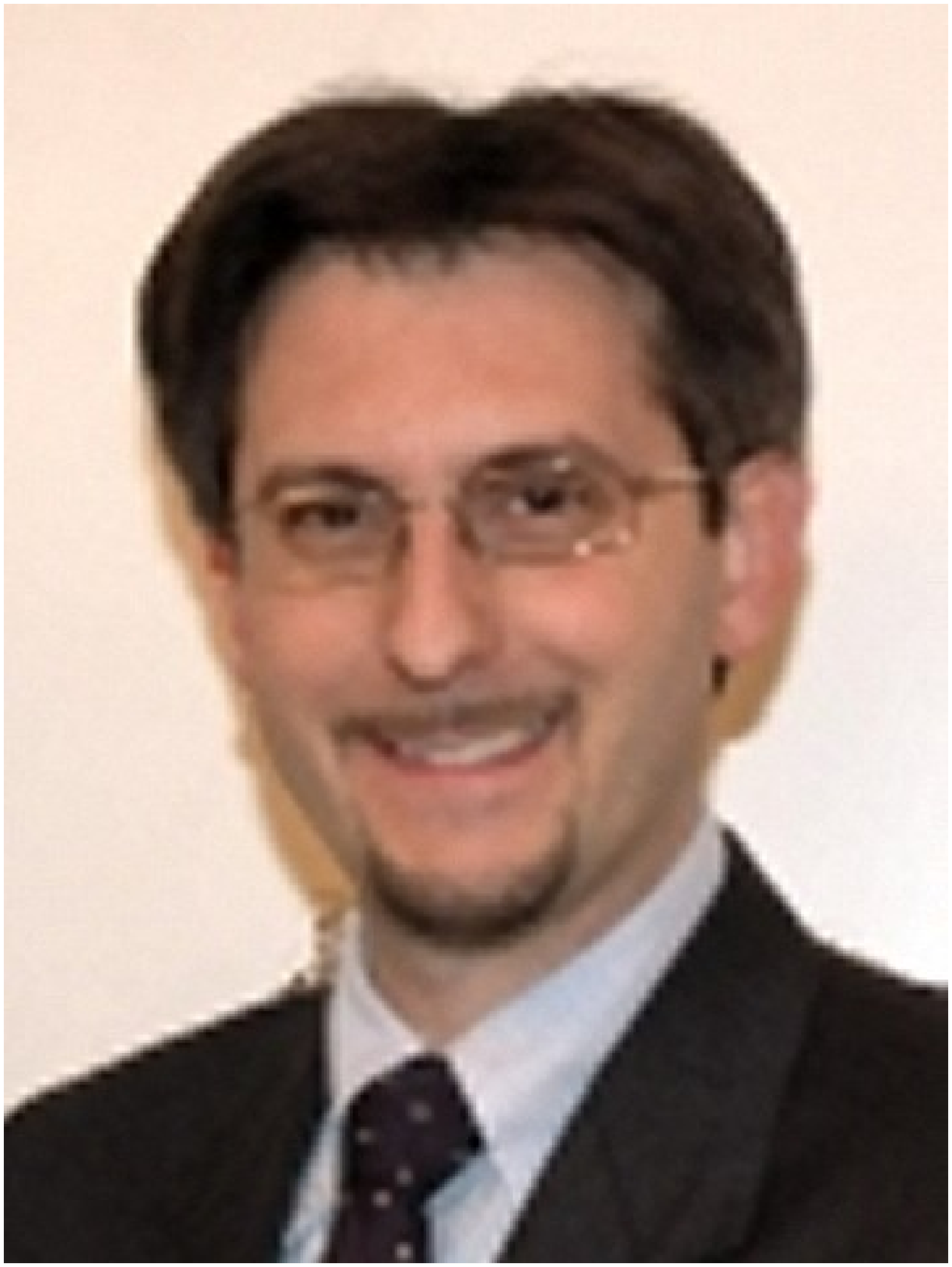}}]{Giulio
    Rosati} received the M.S. degree with honors in Mechanical Engineering from the University of
  Padua, Italy (1999) and the Ph.D. degree in Machine Mechanics from the University of Brescia,
  Italy (2003). He is Associate Professor (since 2006) and was Assistant Professor (2004-2006) of
  Machine Mechanics at the University of Padua, Italy. He was Visiting Professor at the University
  of California at Irvine in 2007.  His research interests include haptics, robotics, industrial
  automation and medical and rehabilitation robotics.  Dr. Rosati is member of the Faculty Board of
  the Doctoral School in Mechatronics Engineering of the University of Padua, Italy (since 2007).
\end{biography}

\end{document}